\documentclass{article}

\usepackage[utf8]{inputenc}
\usepackage{microtype}
\usepackage{graphicx}
\usepackage{subcaption}
\usepackage{booktabs} % for professional tables
\usepackage[table]{xcolor} % 用于表格颜色
\usepackage{multirow}  % 如果需要合并单元格（本例主要用 multicolumn）
\usepackage{algorithmic}

\definecolor{ourbg}{RGB}{236, 242, 255}

\usepackage{listings}
\usepackage[dvipsnames]{xcolor} % 使用标准色卡

\lstdefinestyle{icmlstyle}{
    language=Python,
    basicstyle=\ttfamily\small,      % 字号稍微大一点点，或者用 \footnotesize
    commentstyle=\color{gray}\itshape, % 注释：灰色+斜体
    keywordstyle=\bfseries\color{black}, % 关键字：黑色+加粗 (最正式)
    stringstyle=\color{black},       % 字符串：黑色 (保持整洁)
    numberstyle=\tiny\color{gray},   % 行号：灰色
    breakatwhitespace=false,         
    breaklines=true,                 
    captionpos=t,                    
    keepspaces=true,                 
    numbers=left,                    
    numbersep=8pt,                  
    showspaces=false,                
    showstringspaces=false,
    showtabs=false,                  
    tabsize=4,
    frame=tb,                        % 关键：只显示上下边框 (Top & Bottom)，类似表格
    rulecolor=\color{black},
    framexleftmargin=0em,            % 调整边距使其与正文对齐
    xleftmargin=1em,                 % 代码整体缩进一点，避免行号突兀
    aboveskip=1.5em,                 % 代码块上下的间距
    belowskip=1em
}

\usepackage[most]{tcolorbox}
\usepackage{fontawesome5} % 用于显示漂亮的图标，如大脑、终端等

\newtcolorbox{casestudybox}[2][]{
    enhanced,
    breakable, % 允许跨页
    colback=white,
    colframe=black!70, % 边框颜色
    title={#2}, % 标题
    fonttitle=\bfseries\large,
    attach boxed title to top left={yshift=-2mm, xshift=2mm},
    boxed title style={colback=black!70, sharp corners},
    #1
}

\newcommand{\stepheader}[1]{\vspace{1em}\noindent\textbf{\textsf{#1}}}

\newtcolorbox{thinkbox}{
    enhanced,
    colback=gray!10,
    colframe=gray!30,
    leftrule=3mm,
    arc=0mm,
    boxrule=0pt,
    title={\small \faBrain\ \textit{Thinking Process}},
    coltitle=black!70,
    fonttitle=\bfseries,
    attach boxed title to top left={xshift=3mm, yshift*=-\tcboxedtitleheight/2},
    boxed title style={colback=gray!10, frame hidden},
    top=1em
}

\newtcolorbox{interpreterbox}{
    enhanced,
    colback=black!5,
    colframe=black!50,
    title={\small \faTerminal\ \textit{Interpreter Output}},
    fonttitle=\bfseries\footnotesize,
    coltitle=black,
    attach boxed title to top left={xshift=3mm, yshift*=-\tcboxedtitleheight/2},
    boxed title style={colback=white, frame hidden},
    fontupper=\ttfamily\small,
    top=0.8em
}

\usepackage[T1]{fontenc} % 确保 < > 显示正确

\usepackage{xcolor}

\definecolor{algPhase}{RGB}{0,92,170}   % 蓝：阶段/流程
\definecolor{algPert}{RGB}{160,0,0}     % 红：扰动/失败注入
\definecolor{algOk}{RGB}{0,120,60}      % 绿：正常路径

\newcommand{\algc}[2]{\textcolor{#1}{\itshape // #2}}

\usepackage{hyperref}

\usepackage[preprint]{icml2026}

\usepackage{amsmath}
\usepackage{amssymb}
\usepackage{mathtools}
\usepackage{amsthm}

\newtheoremstyle{schemastyle} % 自定义样式名
  {6pt}   % 上间距
  {6pt}   % 下间距
  {\normalfont} % 也就是由正文部分字体 (roman)，不像 theorem 那样是斜体
  {}      % 缩进量
  {\bfseries} % 标题字体 (Bold)
  {.}     % 标题后的标点
  {0.5em} % 标题后的间距
  {}      % 头部说明 (默认)

\theoremstyle{schemastyle}
\newtheorem{schema}{Schema}

\newcommand{\inlineicon}[1]{%
    \raisebox{-0.15\baselineskip}{%
        \includegraphics[height=1.0\baselineskip]{#1}%
    }%
}

\usepackage[capitalize,noabbrev]{cleveref}

\theoremstyle{plain}

\theoremstyle{definition}

\theoremstyle{remark}

\usepackage[textsize=tiny]{todonotes}

\icmltitlerunning{\textit{DeepTool}: Scaling Interleaved Deliberation in Tool-Integrated Reasoning via Process-Supervised Reinforcement Learning}

\begin{document}

\twocolumn[
  \icmltitle{\textit{DeepTool}: Scaling Interleaved Deliberation in Tool-Integrated Reasoning via Process-Supervised Reinforcement Learning}



  \icmlsetsymbol{equal}{*}

  % 2. 作者列表与单位映射
  \begin{icmlauthorlist}
    \icmlauthor{Yang He}{hit}
    \icmlauthor{Xiao Ding}{hit}
    \icmlauthor{Bibo Cai}{hit}
    \icmlauthor{Yufei Zhang}{hit}
    \icmlauthor{Kai Xiong}{hit}
    \icmlauthor{Zhouhao Sun}{hit}
    \icmlauthor{Bing Qin}{hit}
    \icmlauthor{Ting Liu}{hit}
  \end{icmlauthorlist}

  % 3. 单位详细信息
  \icmlaffiliation{hit}{Research Center for Social Computing and Interactive Robotics, Harbin Institute of Technology, China}

  % 4. 通讯作者（通常为实验室导师，这里以丁效老师为例，可根据实际情况精简或修改）
  \icmlcorrespondingauthor{Xiao Ding}{xding@ir.hit.edu.cn}


  \vskip 0.3in
]



\printAffiliationsAndNotice{}  % no special notice (required even if empty)

\begin{abstract}

Tool-Integrated Reasoning (TIR) extends LLM capabilities by leveraging external environments. However, existing methods lack the deliberation during sequential tool invocation required for strategic planning and self-correction. While RL mitigates this, conventional approaches for Tool-Integrated Reasoning are hindered by sparse outcome-based rewards, failing to supervise \mbox{intermediate} reasoning steps and tool invocations. To address this, we propose \textit{\textbf{DeepTool}}, a novel framework that scales deliberate thinking within the interleaved process of thinking, action, and observation at each turn.
In DeepTool, we first introduce a synthesis pipeline that evolves extended thinking into interleaved trajectories, integrating adversarial perturbations to ensure \mbox{robustness} and self-correction. 
Secondly, we devise \textbf{Process-Supervised Reinforcement Learning} based on GRPO, which utilizes an \textit{Action-Centric Process Reward} to reinforce intermediate interleaved thinking and enforce precise tool invocation at every turn. Extensive experiments demonstrate that DeepTool achieves superior performance, boosting Qwen2.5-7B significantly across six benchmarks (e.g., AIME24: 3.2\% $\to$ 40.4\% and HMMT25: 0.0\% $\to$ 28.6\%). Furthermore, the token cost-effectiveness analysis confirms the utility of interleaved thinking, demonstrating DeepTool's optimal balance between performance and token efficiency. 

% Tool-Integrated Reasoning (TIR) extends LLM capabilities by leveraging external environments. However, existing methods lack the deliberation during sequential tool invocation required for strategic planning and self-correction. While RL mitigates this, conventional approaches for Tool-Integrated Reasoning are hindered by sparse outcome-based rewards, failing to supervise \mbox{intermediate} reasoning steps and tool invocations. To address this, we propose \textit{\textbf{DeepTool}}, a novel framework that scales deliberate thinking within the interleaved process of thinking, action, and observation.
% In DeepTool, we first introduce a synthesis pipeline that evolves extended thinking into interleaved trajectories, integrating adversarial perturbations to ensure \mbox{robustness} and self-correction. 
% Secondly, we devise \textbf{Process-Supervised Reinforcement Learning} based on GRPO, which utilizes an Action-Centric Process Reward to reinforce intermediate interleaved thinking and enforce precise tool invocation. Extensive experiments demonstrate that DeepTool achieves superior performance, boosting Qwen2.5-7B significantly across six benchmarks (e.g., AIME24: 3.2\% $\to$ 40.4\% and HMMT25: 0.0\% $\to$ 28.6\%). Furthermore, cost-effectiveness analysis confirms the utility of interleaved thinking, demonstrating DeepTool's optimal balance between performance and efficiency. 
% Anonymous code and data: \url{https://anonymous.4open.science/r/DeepTool-39B1}.

\end{abstract}

\section{Introduction}
\label{sec:introduction}

Large language models have recently demonstrated substantial progress across a diverse range of reasoning tasks, including mathematical reasoning~\cite{shao2024deepseekmath, ahn2024large, li2024mugglemath}, logical puzzles~\cite{di2025lorp, xie2025logic} and code generation~\cite{zhong2024debug, li2025reasoning}. Despite these advances, purely text-based LLMs are fundamentally limited by their probabilistic nature. They frequently suffer from hallucinations~\cite{zhang2025siren, dhuliawala2024chain}, struggle with tasks requiring precise logic or computation, lack access to real-time information, and are unable to perform actions beyond language generation. To mitigate these inherent shortcomings, tool-integrated reasoning (TIR) has emerged as a promising paradigm~\cite{mialonaugmented, qu2025tool}. By empowering LLMs with diverse external utilities, ranging from code interpreters for exact execution~\cite{feng2025retool, xue2025simpletir} and search engines for up-to-date retrieval~\cite{su-etal-2024-dragin, gao2023retrieval} to multimodal encoders for perceptual tasks~\cite{shieagle, wu2024next}, this approach significantly extends the boundaries and reliability of foundation models~\cite{chenprogram, chen2025toward, qintoolllm}.

%%%%%%%%%%
% Existing approaches for TIR generally fall into two paradigms: supervised fine-tuning (SFT) and reinforcement learning (RL). SFT methods instruct models to invoke tools by imitating curated trajectories~\cite{tang2023toolalpaca, li2023api}, while RL approaches encourage models to interleave reasoning with tool execution, exploring diverse paths optimized by outcome-based rewards~\cite{feng2025retool,li2025torl}. Despite their progress, both paradigms face significant limitations. First, SFT models often suffer from poor generalization~\cite{foster2024behavior, kirkunderstanding}. Since they rely heavily on imitating static examples, they struggle to adapt to scenarios beyond their training distribution~\cite{hao2025understanding, dong2024abilities}. Second, existing frameworks tend to induce ``shallow thinking'', constraining the model’s capacity to internalize sophisticated tool-use strategies~\cite{ghoshal2026tools, qin2025incentivizing}. This limitation largely stems from the suboptimal allocation of computation budgets, which restricts the depth required for complex planning~\cite{xie2024travelplanner, yao2023tree}. Finally, current methods frequently lack logical coherence and dynamic self-correction mechanisms~\cite{huanglarge, shinn2023reflexion, kumartraining}. Without these capabilities, models are prone to error propagation during multi-step reasoning, failing to adjust their strategies adaptively.

Existing approaches for TIR generally fall into two paradigms: supervised fine-tuning (SFT) and reinforcement learning (RL). SFT methods instruct models to invoke tools by imitating curated trajectories~\cite{tang2023toolalpaca, li2023api}. However, since they rely heavily on static examples, they often struggle to generalize beyond their training distribution~\cite{hao2025understanding, dong2024abilities, foster2024behavior, kirkunderstanding}.
While RL approaches mitigate this by encouraging the exploration of diverse paths~\cite{feng2025retool,li2025torl}, they still face significant challenges in maintaining logical coherence and robust error recovery mechanisms~\cite{shinn2023reflexion, kumartraining, huanglarge}, primarily due to the scarcity of gold supervision signals for intermediate reasoning and tool invocation processes. We posit that addressing the aspect of computation budget allocation can markedly enhance their efficacy. Existing frameworks tend to apply uniform ``shallow thinking'' across all steps~\cite{ghoshal2026tools, qin2025incentivizing}. Consequently, they fail to engage in deep deliberation specifically during complex tool invocations~\cite{xie2024travelplanner, yao2023tree}. This deficiency restricts the model's capacity to internalize sophisticated tool-use strategies, leaving them prone to cascading failures when initial actions yield ambiguous or erroneous outputs.

To address these limitations, we introduce \textbf{\textit{DeepTool}}, a framework that reformulates tool-integrated reasoning by scaling \textit{System 2} deliberation~\cite{de2025defining, li2025system} within the interleaved cycle of thinking, acting, and observing. Unlike prior approaches that treat tool invocation as a mere atomic action, \textbf{\textit{DeepTool}} posits that effective tool use requires a continuous, self-correcting cognitive process. To realize this, we first propose \textbf{MOSAIC}, a synthesis pipeline that leverages the System 2 extended thinking of reasoning models to synthesize trajectories enriched with interleaved deliberation. By systematically injecting stochastic adversarial perturbations, MOSAIC compels the model to learn adaptive error recovery and strategic planning. Subsequently, to mitigate the sparsity of outcome-based rewards and the lack of supervision for intermediate reasoning steps and tool invocations, we devise a \textbf{Process-Supervised Reinforcement Learning} strategy based on GRPO. Leveraging a novel \textit{Action-Centric Process Reward}, this mechanism provides dense supervision at every turn, thereby further scaling interleaved deliberation during reasoning while reinforcing precise tool manipulation. 
The contributions of this paper are summarized as follows:
\begin{itemize}
    \item We propose \textit{\textbf{DeepTool}}, a novel framework that scales System 2 deliberation within the interleaved process of thinking, action, and observation at each turn, enabling LLMs to perform strategic planning and self-correction during multi-turn environmental interactions.
    \item We introduce \textbf{MOSAIC}, a synthesis pipeline that incorporates interleaved deliberation and injects adversarial perturbations during training data generation, significantly enhancing the model's robustness and error-recovery capabilities.
    \item We devise \textbf{Process-Supervised Reinforcement Learning} with Action-Centric Process Rewards that provides dense supervision to further scale interleaved deliberation and enforce precise tool invocations.
    \item Extensive experiments across diverse challenging benchmarks demonstrate the superiority of \textit{\textbf{DeepTool}}, while cost-effectiveness analysis confirms that interleaved thinking strikes an optimal balance between performance and token efficiency.
\end{itemize}

\section{Methodology}
\label{sec:methodology}

%%%%

We propose \textbf{\textit{DeepTool}}, a framework that scales deliberate thinking within the interleaved process of thinking, action, and observation, explicitly empowering models to perform strategic planning and self-correction. Our approach proceeds in three stages:

\begin{enumerate}
    \item \textbf{Trajectory Synthesis (\S\ref{sec:mosaic_synthesis}):} We introduce \textbf{MOSAIC} to generate interleaved thinking trajectories, injecting adversarial perturbations to enforce robustness against execution failures.
    \item \textbf{Behavioral Initialization (\S\ref{sec:cold_start_sft}):} We apply cold-start fine-tuning to align the model with the protocol of alternating between deliberate thinking, tool execution, and environmental observation.
    \item \textbf{Process Refinement (\S\ref{sec:psrl}):} We employ process-supervised reinforcement learning to optimize reasoning at the step level, utilizing an Action-Centric Process Reward for dense supervision.
\end{enumerate}

%%%%%%% 长一些版本的version %%%%%%%%%%

% We propose \textbf{\textit{DeepTool}}, a framework that interleaves deliberate thinking with tool invocation, explicitly empowering models to perform strategic planning and self-correction. Our approach proceeds in three stages:

% \begin{enumerate}
%     \item \textbf{Trajectory Synthesis (\S\ref{sec:mosaic_synthesis}):} Introducing \textbf{MOSAIC}, a synthesis pipeline that integrates extended thinking into the tool-use flow to construct trajectories characterized by interleaved thinking, while injecting adversarial perturbations to improve robustness.

%     \item \textbf{Behavioral Initialization (\S\ref{sec:cold_start_sft}):} Performing cold-start fine-tuning to internalize the interleaved thinking format, aligning the model with a structured protocol of alternating between thinking, tool execution, and environmental observation.
    
%     \item \textbf{Process Refinement (\S\ref{sec:psrl}):} Leveraging step-wise decomposition to factorize reasoning trajectories, enabling process-supervised reinforcement learning to optimize the model at the granularity of individual steps via Action-Centric Process Reward.
% \end{enumerate}

%%%%%%%%%%%%%%%%%%%%%%%%%%%%%%%%%%%%

% ----------------------------------------------------------------------
% Section Structure for MOSAIC
% ----------------------------------------------------------------------

\subsection{MOSAIC: Synthesizing Self-Correcting Interleaved Trajectories}
\label{sec:mosaic_synthesis}

% 【排版技巧】将图片代码放在章节开头，利用 [t] 参数让其浮动到本页或下一页顶部
% 这样不会打断读者的文本阅读流
\begin{figure}[t]
  % \vskip 0.2in
  \begin{center}
    \centerline{\includegraphics[width=\columnwidth]{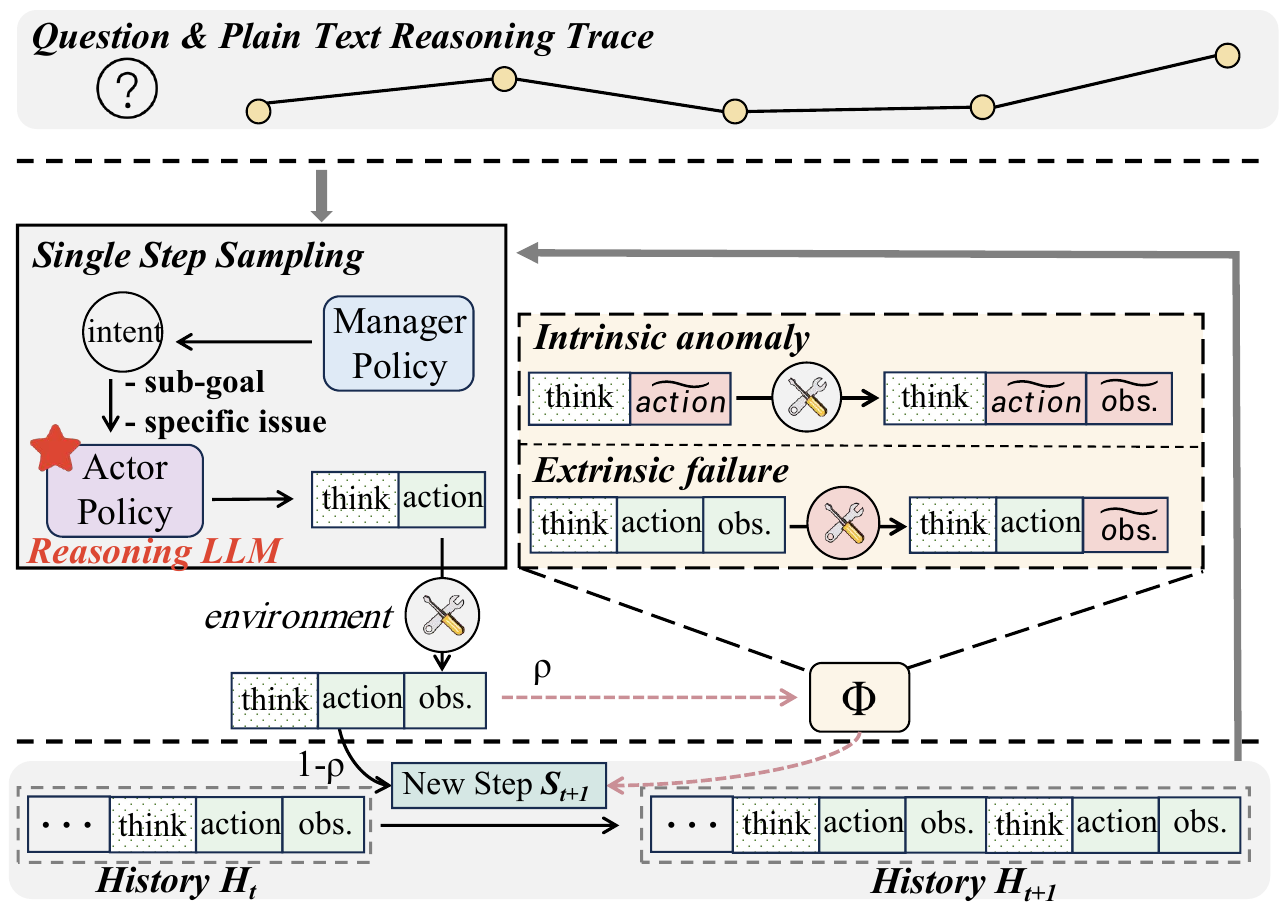}}
    \caption{
      The \textbf{MOSAIC} synthesis pipeline. 
      A hierarchical Manager-Actor architecture is employed, utilizing a \textbf{reasoning model} as the Actor Policy (\inlineicon{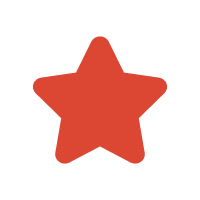}) to synthesize \textbf{System 2 level deliberation} (\inlineicon{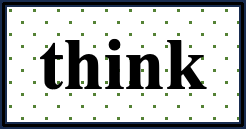}) based on strategic intents.
      The operator $\Phi$ stochastically injects perturbations (with probability $\rho$) to create \textit{perturbed steps}, fostering robust error recovery.
    }
    \label{fig:mosaic_pipeline}
  \end{center}
  \vskip -0.2in
\end{figure}

% 开篇介绍，引用图片
% We introduce MOSAIC (\textbf{M}odel-\textbf{O}rchestrated \textbf{S}ynthesis via \textbf{A}dversarial \textbf{I}njection \& \textbf{C}orrection), a framework that transforms static, monolithic \textit{Extended Thinking} into flexible \textit{Interleaved Thinking}. By leveraging stochastic adversarial perturbations, MOSAIC evolves rigid reasoning chains into robust, self-correcting Tool-Integrated Reasoning processes, which is shown in Figure~\ref{fig:mosaic_pipeline}.

We introduce MOSAIC (\textbf{M}odel-\textbf{O}rchestrated \textbf{S}ynthesis via \textbf{A}dversarial \textbf{I}njection \& \textbf{C}orrection) to transform \textit{Extended Thinking} into \textit{Interleaved Thinking} within Tool-Integrated Reasoning. As illustrated in Figure~\ref{fig:mosaic_pipeline}, by leveraging stochastic adversarial perturbations, MOSAIC evolves rigid reasoning chains into robust, self-correcting trajectories.

% \subsubsection{Formulation and Methodology}

% 问题定义
\paragraph{Problem Definition.}
Given an input query $x$ and a plain text reasoning trace $y^*$, our objective is to synthesize a trajectory $H_t = [s_1, \dots, s_t]$, where each step $s_t = (th_t, a_t, o_t)$ consists of a deliberate thinking $th_t$, an executable action $a_t \in \mathcal{A}$ (where the action space $\mathcal{A}$ encompasses both intermediate code snippets $c_t$ and the terminal answer $ans$), and an environmental observation $o_t$.

% 核心架构
\paragraph{Hierarchical Synthesis Architecture.}
We employ a bilevel policy framework to generate the next step $s_{t+1}$ given context $(x, y^*, H_t)$:
\begin{itemize}
    \item \textbf{Manager ($\pi_{\mathcal{M}}$):} A high-level planner that synthesizes a strategic intent $I_{t+1}$ (e.g., sub-goal formulation or error rectification):
    \begin{equation}
        I_{t+1} \sim \pi_{\mathcal{M}}(\cdot \mid x, y^*, H_t)
    \end{equation}
    \item \textbf{Actor ($\pi_{\mathcal{A}}$):} A low-level executor that materializes the intent into specific reasoning and action:
    \begin{equation}
        (th_{t+1}, \hat{a}_{t+1}) \sim \pi_{\mathcal{A}}(\cdot \mid x, H_t, I_{t+1})
    \end{equation}
\end{itemize}

% 扰动机制
\paragraph{Adversarial Perturbation Mechanism.}
To induce robustness against failure, we introduce a stochastic perturbation operator $\Phi$ that intercepts the actor's output $\hat{a}_t$ with probability $\rho$. The synthesis procedure is formally detailed in Algorithm~\ref{alg:mosaic_synthesis}. The operator bifurcates into two distinct failure modes:

\begin{enumerate}
    \item \textbf{Intrinsic Generative Anomaly ($\Phi_{gen}$):} Simulates model-side defects (e.g., syntax errors). The action is mutated into a corrupted version $\tilde{a}_t$, yielding a traceback observation:
    \begin{equation}
        s_t = (th_t, \underbrace{\Phi_{gen}(\hat{a}_t)}_{\tilde{a}_t}, \text{Exec}(\tilde{a}_t))
    \end{equation}
    \item \textbf{Extrinsic Environmental Failure ($\Phi_{env}$):} Simulates system instabilities (e.g., timeouts). The valid action $\hat{a}_t$ is preserved, but the environment is intercepted to return a synthetic error signal:
    \begin{equation}
        s_t = (th_t, \hat{a}_t, \underbrace{\Phi_{env}(\hat{a}_t)}_{\tilde{o}_{err}})
    \end{equation}
\end{enumerate}
In the absence of perturbation, the standard transition applies: $s_t = (th_t, \hat{a}_t, \text{Exec}(\hat{a}_t))$.

\begin{algorithm}[tb]
  \caption{\textbf{MOSAIC} Trajectory Synthesis via Hierarchical Planning and Stochastic Adversarial Injection}
  \label{alg:mosaic_synthesis}
  \begin{algorithmic}
    \STATE \textbf{Input:} query $x$, reference trace $y^*$, Manager $\pi_{\mathcal{M}}$, Actor $\pi_{\mathcal{A}}$, perturb rate $\rho$, max steps $T_{\max}$
    \STATE \textbf{Output:} interleaved trajectory $H$
    \STATE $H \leftarrow \emptyset$
    \FOR{$t=1$ \textbf{to} $T_{\max}$}
      \STATE \algc{algPhase}{Manager: sample strategic intent} \\ 
      $I_t \sim \pi_{\mathcal{M}}(\cdot \mid x, y^*, H)$
      \STATE \algc{algPhase}{Actor: generate thought and candidate action} \\
      $(th_t, \hat{a}_t) \sim \pi_{\mathcal{A}}(\cdot \mid x, H, I_t)$
      
      \IF{$\hat{a}_t = ans$}
        \STATE $H \leftarrow H \cup \{(th_t, ans, \varnothing)\}$; \textbf{break}
      \ENDIF
      
      \IF{$\hat{a}_t$ is code \textbf{and} random() $< \rho$}
        \STATE \algc{algPert}{Adversarial injection: choose failure mode}
        \IF{coin\_flip() is Heads} 
          \STATE \algc{algPert}{Intrinsic anomaly $\Phi_{gen}$: corrupt action, then execute}
          \STATE $a_t \leftarrow \Phi_{gen}(\hat{a}_t)$; \quad $o_t \leftarrow \text{Exec}(a_t)$
        \ELSE
          \STATE \algc{algPert}{Extrinsic failure $\Phi_{env}$: keep action, corrupt observation}
          \STATE $a_t \leftarrow \hat{a}_t$; \quad $o_t \leftarrow \Phi_{env}(\hat{a}_t)$
        \ENDIF
      \ELSE
        \STATE \algc{algOk}{Standard execution}
        \STATE $a_t \leftarrow \hat{a}_t$; \quad $o_t \leftarrow \text{Exec}(a_t)$
      \ENDIF
      
      \STATE $H \leftarrow H \cup \{(th_t, a_t, o_t)\}$
    \ENDFOR
    \STATE \textbf{return} $H$
  \end{algorithmic}
\end{algorithm}

\subsection{Cold-Start Fine-Tuning for Interleaved TIR}
\label{sec:cold_start_sft}

We initialize Tool-Integrated Reasoning via full-parameter supervised fine-tuning on synthesized trajectories.
Each query $x$ is wrapped with a dedicated TIR system prompt that specifies the tool environment and enforces interleaved reasoning (Appendix~\ref{lst:DeepTool-Infer}).

\paragraph{Standardized Interleaved Format.}
We enforce a strict syntax to distinguish between internal thought, tool invocation, and environment observation. Let a training trajectory be denoted as $\tau = [s_1, s_2, \dots, s_T]$. The structure of step $s_t$ adapts dynamically to the action type:

\begin{schema}[Intermediate Tool Step ($a_t = c_t$)]
\label{sch:intermediate}
This step consists of reasoning, code generation, and the interpreter's feedback:
\[
\begin{aligned}
    s_t = \; & \texttt{<think>} \, th_t \, \texttt{</think>} \\
          & \texttt{<code>} \, c_t \, \texttt{</code>} \\
          & \texttt{<interpreter>} \, o_t \, \texttt{</interpreter>}
\end{aligned}
\]
\end{schema}

\begin{schema}[Terminal Answer Step ($a_t = ans$)]
\label{sch:terminal}
The trajectory concludes with a final thought and the answer:
\[
\begin{aligned}
    s_T = \; & \texttt{<think>} \, th_T \, \texttt{</think>} \\
          & \texttt{<answer>} \, ans \, \texttt{</answer>}
\end{aligned}
\]
\end{schema}

This schema enforces unambiguous role separation between thinking, tool invocation, and observed feedback, yielding deterministic execution boundaries that establish a stable initialization critical for subsequent RL training.

%%%%%%%

\subsection{Process-Supervised Reinforcement Learning}
\label{sec:psrl}

\begin{figure*}[t]  % 使用星号环境，[t] 表示放在页面顶部 (top)
  \centering         % 推荐使用 \centering 代替 \begin{center} 以减少多余间距
  \includegraphics[width=0.95\textwidth]{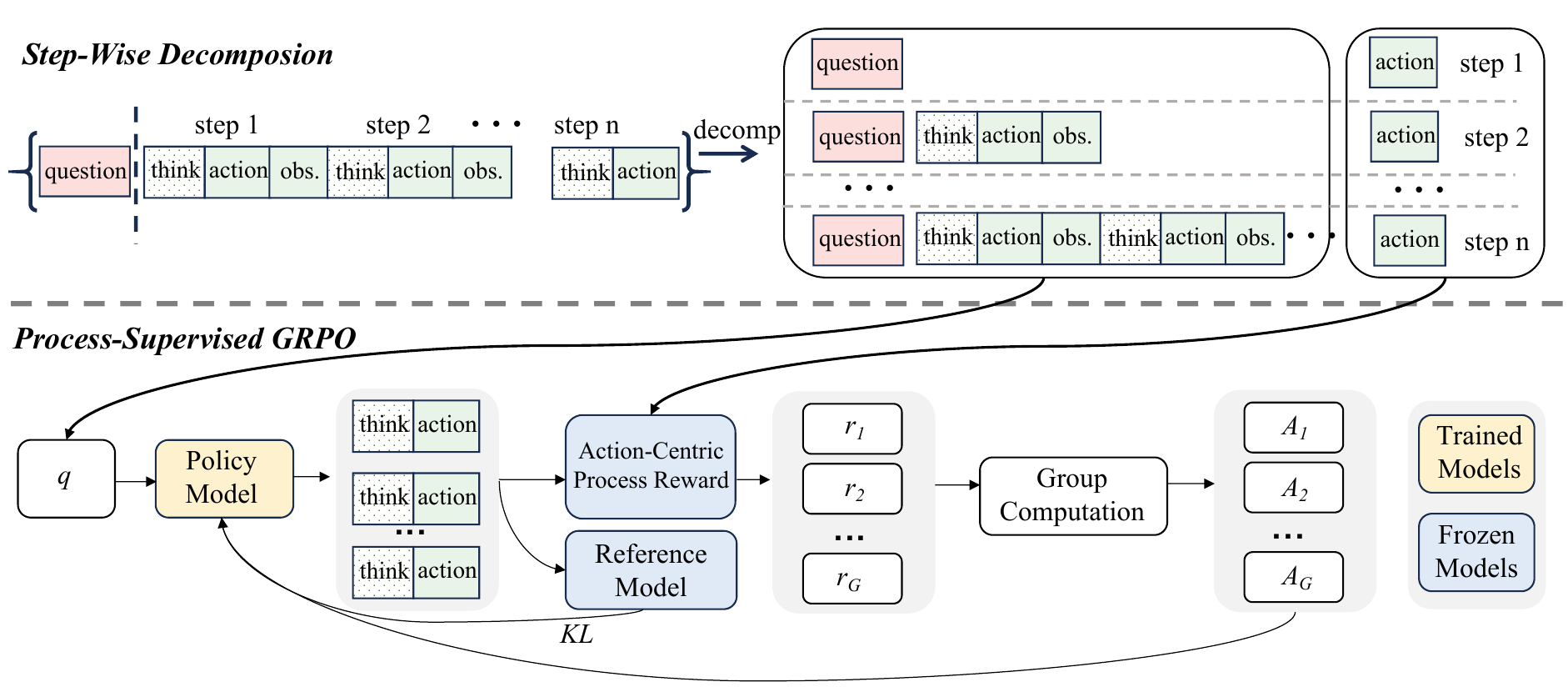} % 跨栏图通常建议设置宽度为 textwidth 的比例
%   \caption{
%   \textbf{Overview of the Process-Supervised Reinforcement Learning framework.} 
%   The pipeline begins with \textbf{Step-Wise Decomposition}, where expert trajectories are factorized into discrete training instances conditioned on ground-truth history ($H^*_{t-1}$) to enable dense supervision. 
%   At each step, the policy samples a group of candidate outputs (thoughts and actions). These candidates are evaluated via the \textbf{Action-Centric Process Reward}, 
%   a hybrid mechanism that strictly enforces format protocols and measures the similarity of the generated action $a_t$ against the gold reference $a^*_t$, while explicitly permitting diverse reasoning pathways in the thought process $th_t$. Finally, the policy is optimized using Group Relative Policy Optimization (GRPO).
% }

\caption{
\textbf{Overview of the Process-Supervised Reinforcement Learning framework.} 
The pipeline starts with \textbf{Step-Wise Decomposition}, factorizing trajectories based on ground-truth history ($H^*_{t-1}$) for dense supervision. 
Rollouts are evaluated via the \textbf{Action-Centric Process Reward}, enforcing action alignment ($a_t$ vs $a^*_t$) while permitting diverse reasoning ($th_t$), optimizing the policy using GRPO.
}

  \label{fig:psrl}
\end{figure*}

\subsubsection{Step-Wise Decomposition}
Outcome-based reinforcement learning inherently suffers from reward sparsity and the credit assignment problem, as it lacks feedback for intermediate reasoning. Moreover, it is vulnerable to error accumulation, where early deviations render subsequent optimization signals ineffective.

To enable dense supervision, we decompose the expert trajectory $\tau^*$ into discrete, independent training instances. For each step $t$, we define the input context $Q_t$ and target $L_t$ as:
\begin{equation}
    Q_t = (x, H^*_{t-1}), \quad L_t = (th^*_t, a^*_t)
\end{equation}
We utilize the ground-truth history $H^*_{t-1}$, to supervise each step independently, effectively neutralizing error propagation. This converts a single trajectory into $n$ dense training signals, as shown in Figure~\ref{fig:psrl}.

\subsubsection{Process-Supervised GRPO}

\paragraph{Action-Centric Process Reward}
To enforce precise execution while preserving reasoning flexibility, we introduce the Action-Centric Process Reward. This step-wise mechanism $r(s_t, s^*_t)$ combines a strict structural gate with a soft similarity metric:
\begin{equation}
    r(s_t, s^*_t) = \mathbb{I}_{\text{fmt}}(th_t, a_t) \cdot r_{\text{sim}}(a_t, a^*_t)
\end{equation}
The binary gate $\mathbb{I}_{\text{fmt}}$ enforces adherence to the phase-specific protocols defined in Schema \ref{sch:intermediate} and Schema \ref{sch:terminal}:
\begin{equation}
    \mathbb{I}_{\text{fmt}} = 
    \begin{cases} 
    1 & \text{if } t < n, \text{ satisfies Schema \ref{sch:intermediate}} \\
    1 & \text{if } t = n, \text{ satisfies Schema \ref{sch:terminal}} \\
    0 & \text{otherwise}
    \end{cases}
\end{equation}
The term $r_{\text{sim}}$ quantifies the alignment of the generated action $a_t$ against the reference $a^*_t$ via Gestalt Pattern Matching. Defined as the normalized measure of recursively identified common subsequences, the metric is given by:
\begin{equation}
    r_{\text{sim}}(a_t, a^*_t) = \frac{2 \cdot \sum_{k \in \text{Matches}} \text{len}(k)}{|a_t| + |a^*_t|}
\end{equation}
where the numerator represents the doubled sum of non-overlapping matching blocks. By restricting evaluation to $a_t$ and excluding the thought process $th_t$, we incentivize the model to explore diverse cognitive pathways provided they converge to the correct executable outcome.

\paragraph{Optimization via GRPO}

%%%%%

\begin{table*}[t] % table* 表示跨双栏，[t] 表示置顶
  \centering
  \small
  \caption{Main results on mathematical reasoning benchmarks. Best results are highlighted in bold. Unless noted, Qwen2.5-7B is used as the backbone. For ToRL and ReTool, their RL models generate training data inferences for SFT, which serves as the baseline.}
  \label{tab:main_results}
  \renewcommand{\arraystretch}{1.2} % 稍微增加行高，防止文字和背景色显得太挤
  
  % 调整列间距，如果表格太宽可以减小这个数值，例如 \setlength{\tabcolsep}{4pt}
  % \setlength{\tabcolsep}{5pt} 

  \begin{tabular}{l ccccccc}
    \toprule
    \textbf{Method} & \textbf{AIME24} & \textbf{AIME25} & \textbf{MATH500} & \textbf{Olympiad} & \textbf{AMC23} & \textbf{HMMT25} & \textbf{GPQA-D} \\
    \midrule
    
    % --- 第二部分：Qwen2.5-7B ---
    \multicolumn{8}{c}{\textit{models based on Qwen2.5-7B}} \\
    \midrule
    Qwen2.5-7B-Base & 3.2 & 1.1 & 51.9 & 15.4 & 21.7 & 0.0 & 32.4 \\
    Qwen2.5-7B-TIR & 1.7 & 0.6 & 18.0 & 6.2 & 10.8 & 1.9 & 28.5 \\
    Search-R1 & 16.7 & 6.7 & 63.8 & - & 45.0 & - & 33.3 \\
    Search-o1 & 6.7 & 10.0 & 61.8 & - & 37.5 & - & - \\
    ToRL & 30.0 & 26.7 & 80.2 & - & 67.5 & - & - \\
    ReTool & 23.3 & \textbf{30.0} & 78.4 & - & 62.5 & - & - \\
    SimpleRL-Zoo-7B & 15.6 & - & 78.2 & 40.4 & 62.5 & - & - \\
    ZeroTIR-7B & 39.6 & 25.0 & 80.2 & - & - & 22.5 & - \\
    ARPO-7B & 30.0 & 30.0 & 78.8 & - & - & - & - \\
    
    % 给 ours 的行添加背景色
    \rowcolor{ourbg} \textbf{\textit{DeepTool (w/o thinking)}} & 15.0 & 14.2 & 65.6 & 27.4 & 45.6 & 10.0 & 43.1 \\
    \rowcolor{ourbg} \textbf{\textit{DeepTool (SFT)}} & 38.3 & 25.8 & 70.4 & 38.5 & \textbf{68.1} & 20.4 & 44.8 \\
    \rowcolor{ourbg} \textbf{\textit{DeepTool (+RL)}} & \textbf{40.4} & 28.6 & \textbf{84.7} & \textbf{48.8} & 64.0 & \textbf{28.6} & \textbf{45.3} \\

    \midrule

    % --- 第一部分：Qwen3-4B ---
    \multicolumn{8}{c}{\textit{models based on Qwen3-4B}} \\
    \midrule
    Qwen3-4B-Base & 7.1 & 6.3 & 70.1 & 33.4 & 48.1 & 2.5 & 32.6 \\
    
    % 给 ours 的行添加背景色
    \rowcolor{ourbg} \textbf{\textit{DeepTool (w/o thinking)}} & 17.1 & 13.3 & 69.4 & 31.6 & 50.9 & 11.3 & 34.5 \\
    \rowcolor{ourbg} \textbf{\textit{DeepTool (SFT)}} & 36.2 & \textbf{35.0} & 82.7 & 39.2 & 74.4 & 20.0 & 37.4 \\
    \rowcolor{ourbg} \textbf{\textit{DeepTool (+RL)}} & \textbf{40.0} & 30.0 & \textbf{83.6} & \textbf{49.8} & \textbf{75.3} & \textbf{23.6} & \textbf{43.8} \\
    
    \bottomrule
  \end{tabular}
\end{table*}

%%%%%

For each target step $t$, we treat the model output as a single sequence
$o_{i} \triangleq (th_{i,t}, a_{i,t})$ (i.e., the concatenation of \textit{think} and \textit{action}),
and define the conditioning query as $q \triangleq (x, H^*_{t-1})$.
We sample a group of candidates $\{o_i\}_{i=1}^G \sim \pi_{\theta_{\text{old}}}(\cdot \mid q)$ and compute
their action-centric process rewards $r_{i}=r(o_{i}, s_t^*)$. The group-relative advantage is normalized as
\begin{equation}
\hat{A}_{i,k}=\frac{r_{i}-\mathrm{mean}(\mathbf{r})}{\mathrm{std}(\mathbf{r})}, \quad
\mathbf{r}=\{r_{i}\}_{i=1}^G .
\end{equation}
Let $o_{i,k}$ denote the $k$-th token of $o_i$ and $o_{i,<k}$ its prefix. The probability ratio is
\begin{equation}
\rho_{i,k}=\frac{\pi_\theta\!\left(o_{i,k}\mid q, o_{i,<k}\right)}
{\pi_{\theta_{\text{old}}}\!\left(o_{i,k}\mid q, o_{i,<k}\right)} .
\end{equation}
Following the clipped surrogate objective, the GRPO loss for step $t$ is defined as follows (see Section \ref{sec:impl_details} for details):

\begin{equation}
\label{eq:grpo}
\begin{split}
& \mathcal{L}_{\text{GRPO}}(\theta) = -\frac{1}{\sum_{i=1}^G |o_i|}\sum_{i=1}^G \sum_{k=1}^{|o_i|} \Bigg[ \\
& \quad \min\Big(\rho_{i,k}\hat{A}_{i,k},\, \mathrm{clip}(\rho_{i,k},1-\epsilon,1+\epsilon)\hat{A}_{i,k}\Big) \\
& \quad -\beta\,\mathbb{D}_{\mathrm{KL}}\!\left[\pi_\theta\|\pi_{\mathrm{ref}}\right] \Bigg].
\end{split}
\end{equation}

This formulation directly optimizes the joint \textit{think+action} output $o$ under history context $q_t$,
stabilizing training; since the soft similarity reward is computed only on $a_t$, diverse thoughts are permitted
as long as they produce better actions with higher relative rewards.

%%%%%%%%%%%%%%%%%%%%%%%%%%%%

\section{Experiment}
\subsection{Setup}
\label{subsec:setup}

\paragraph{Training}
We evaluate Qwen2.5-7B-Instruct~\cite{qwen2025qwen25technicalreport} and Qwen3-4B-Base~\cite{yang2025qwen3} across three configurations: a \textit{w/o thinking} SFT baseline, a \textit{interleaved thinking} SFT model, and its further optimization via process-supervised Reinforcement Learning (RL). The SFT stage samples 8k instances from OpenR1-Math-220k~\cite{openr1_math_220k} with a Mosaic parameter of $\rho=0.1$, while the RL stage employs the LIMO dataset~\cite{ye2025limo} with $\rho=0.2$. Detailed hyperparameter configurations for the SFT and RL stages, as well as the specifications for data synthesis, are provided in Appendix~\ref{sec:impl_details}.

\paragraph{Evaluation}
We evaluate our method on a comprehensive set of mathematical reasoning benchmarks, including MATH500~\cite{hendrycks2measuring}, AIME 2024, AIME 2025, AMC 23, HMMT Feb 2025, OlympiadBench~\cite{he2024olympiadbench}, and GPQA-Diamond~\cite{rein2024gpqa}. We report the average@8 scores to minimize variance. For comparison, we categorize baselines into three groups: search-based approaches, such as Search-R1~\cite{jin2025searchr} and Search-o1~\cite{li-etal-2025-search}; RL-enhanced tool reasoning methods, including ToRL~\cite{li2025torl}, ReTool~\cite{feng2025retool}, and ARPO~\cite{dong2025agentic}; and Zero-RL paradigms like SimpleRL-Zoo~\cite{zeng2025simplerl} and ZeroTIR~\cite{mai2025agent}. 
% We note that a portion of the baseline statistics is sourced from prior literature \citep{chen2025toward}.

\subsection{Main Results}

Table~\ref{tab:main_results} reports the main results on a diverse suite of mathematical reasoning benchmarks. Overall, \textit{\textbf{DeepTool}} yields strong and consistent improvements across both backbones, and achieves the best performance on the majority of evaluated datasets.

\paragraph{Overall performance.}
\textit{\textbf{DeepTool}} delivers substantial gains over the corresponding base models on nearly all benchmarks. With Qwen2.5-7B, it reaches 40.4 on AIME24 and 84.7 on MATH500, and further sets the strongest results on challenging benchmarks such as OlympiadBench and GPQA-D. The same trend holds for the smaller Qwen3-4B backbone, where DeepTool remains highly competitive on key competition-style and long-horizon reasoning evaluations, demonstrating strong robustness to model scale and backbone choice.

\paragraph{Effect of two-stage training.}
We observe \textit{monotonic} improvements along \textit{\textbf{DeepTool}} two-stage pipeline. Moving from \textit{DeepTool (w/o thinking)} to \textit{DeepTool (SFT)} consistently improves performance, with particularly clear benefits on competition-oriented benchmarks (e.g., AIME/AMC). Further applying process-supervised RL (\textit{DeepTool (+RL)}) brings additional gains across benchmarks, such as MATH500 and OlympiadBench. These results suggest that SFT and RL provide complementary benefits: SFT stabilizes tool-augmented reasoning behaviors, while RL strengthens step-wise decision quality under more challenging regimes.

\paragraph{Ablation on interleaved thinking.}
The \textit{w/o thinking} setting isolates the role of explicit interleaved thinking and shows clear degradation across datasets and backbones. In contrast, \textit{interleaved thinking} in \textit{DeepTool (SFT)} consistently recovers large improvements, indicating that training with interleaved thinking is a key driver of \textit{\textbf{DeepTool}}'s effectiveness. This ablation underscores that the gains are not solely from tool access, but from learning to coordinate tool use with structured intermediate thinking. For a concrete qualitative comparison demonstrating how \textit{\textbf{DeepTool}} recovers from errors where the non-thinking baseline fails, please refer to the \textbf{detailed case studies} in Appendix~\ref{appendix:casestudy}.

\subsection{State Preservation in Interleaved Thinking}
\label{subsec:preserve_state}

\begin{figure}[t]
  % \vskip 0.2in
  \begin{center}
    \centerline{\includegraphics[width=\columnwidth]{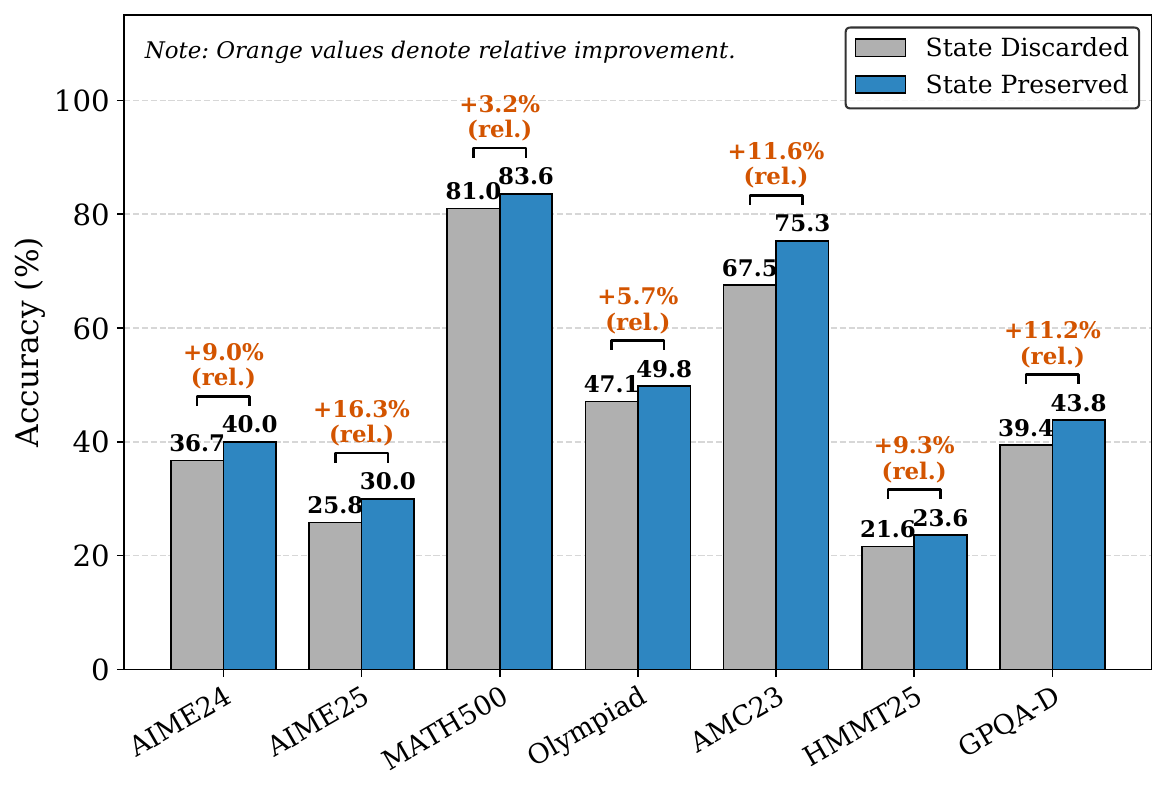}}
    \caption{Performance comparison between discarding and preserving interleaved reasoning state. Orange labels indicate the relative improvement brought by state preservation. The results demonstrate that maintaining a continuous thinking state significantly enhances reliability across all benchmarks.}
    \label{fig:state_ablation}
  \end{center}
  \vskip -0.2in
\end{figure}

\textit{\textbf{DeepTool}} is inherently multi-turn: the model repeatedly \emph{thinks}, \emph{acts} via tools, and \emph{observes} the returned environment feedback. A central design choice is whether to carry forward the intermediate thinking trace (the ``thinking state'') across turns, or to discard it and propagate only tool outputs. We study this with two variants, \textit{DeepTool (state discarded)} and \textit{DeepTool (state preserved)}, and summarize the comparison in Figure~\ref{fig:state_ablation}.

Preserving the interleaved thinking state consistently improves performance on all seven benchmarks. The gains are most pronounced on high-difficulty math: on AIME 2025, accuracy rises from 25.8\% to 30.0\% (a +16.3\% relative improvement), and AIME 2024 increases by +9.0\% relatively, and similarly on AMC 2023 and GPQA-Diamond (+11.6\% and +11.2\% relative gains). Even on stronger-baseline settings like MATH500, it provides a stable improvement (+3.2\%), indicating robustness across difficulty regimes.

\paragraph{Analysis}
We attribute these improvements to two complementary effects:
\begin{itemize}
    \item \textbf{Mitigating State Drift:} State preservation ensures that accumulated constraints, hypotheses, and partial conclusions remain available, preventing the need to repeatedly reconstruct context from tool outputs.
    \item \textbf{Promoting Loop Stabilization:} State preservation stabilizes the \emph{thinking--acting--observation} loop, enabling iterative refinement and targeted self-correction over long horizons.
\end{itemize}
In contrast, discarding state fragments the TIR working context, increases redundant reasoning, and amplifies compounding errors in tool-heavy trajectories, especially in multi-step solve-and-verify workflows.

\subsection{Scaling Interleaved Thinking Budget}
\label{subsec:thinking_budget}

\begin{figure}[t]
  % \vskip 0.2in
  \begin{center}
    \centerline{\includegraphics[width=\columnwidth]{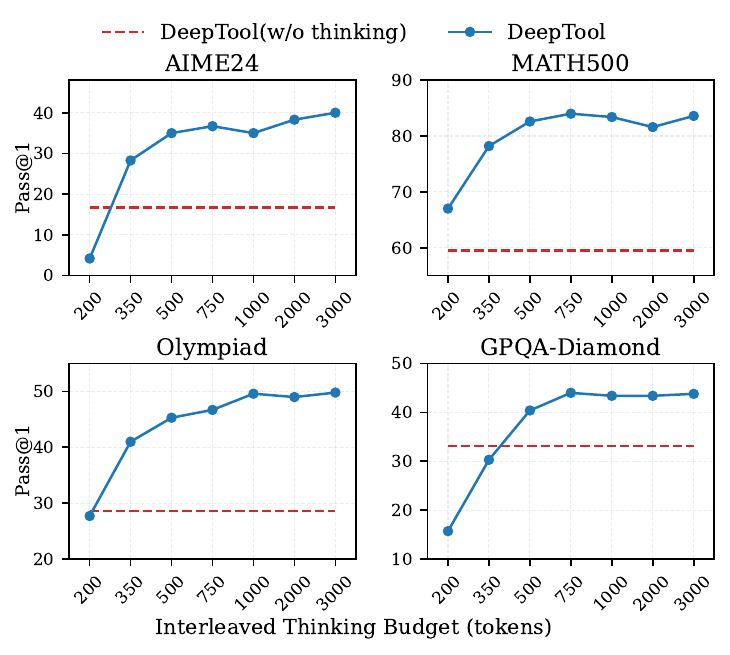}}
    \caption{
      Performance scaling with respect to the per-step thinking token budget. The red dashed line represents the baseline \textit{DeepTool (w/o thinking)}. Results indicate that sufficient thinking budget is crucial.
    }
    \label{fig:thinking_scaling}
  \end{center}
  \vskip -0.2in
\end{figure}

In the Interleaved Thinking setting, the model is given a per-step token budget to produce an explicit thinking process. We sweep this budget and compare against the \textit{DeepTool (w/o thinking)} baseline (red dashed line in Figure~\ref{fig:thinking_scaling}). Full evaluation results across all tested benchmarks are provided in Appendix Figure~\ref{fig:budget_full}. % 或者引用附录的章节标签

\paragraph{Analysis of scaling effects.}
Increasing the thinking budget consistently improves performance across benchmarks, yielding a smooth upward trend. Additional reasoning space supports more complete planning and self-checking, which translates into higher tool-use accuracy. The gains gradually taper at larger budgets, indicating diminishing returns once the model has sufficient budget to reason.

\paragraph{Budget needs depend on task difficulty.}
When the budget is too small, explicit interleaved reasoning can be ineffective and may even fall below the non-thinking baseline, especially on harder benchmarks, because the reasoning trace becomes shallow or truncated. Easier datasets reach the flat region earlier, while harder datasets continue benefiting from larger budgets before saturating. This suggests selecting the thinking budget according to expected difficulty, balancing accuracy against inference cost.

\subsection{Token Cost-Effectiveness Analysis}
\label{subsec:efficiency}

%%%%%%%%

% We evaluate the trade-off between performance improvement and computational overhead by analyzing the \textbf{cost-effectiveness} of Interleaved Thinking (Figure~\ref{fig:efficiency}).
% Let $A_{\text{DeepTool}}$ and $A_{\text{base}}$ denote the accuracies of our method and the \textit{DeepTool (w/o thinking)} baseline, respectively. Let $C_{\text{DeepTool}}$ represent the average token cost per problem for our method. We define the cost-effectiveness metric (y-axis) as:
% \begin{equation}
% \label{eq:efficiency}
% \mathrm{Eff} \;=\; \frac{A_{\text{DeepTool}} - A_{\text{base}}}{C_{\text{DeepTool}}}\times 1000,
% \end{equation}
% which quantifies the accuracy gain achieved per $1,000$ tokens consumed by DeepTool.
% Unlike a raw accuracy comparison, this metric normalizes the performance boost by the computational cost required to achieve it.
% A higher positive $\mathrm{Eff}$ value indicates a high return on computational investment. Conversely, a negative value implies that the method underperforms the baseline despite the token usage. The shaded regions above and below the baseline correspond to the magnitude of net accuracy gain and loss.

%%%%%%%%%

\begin{figure}[t]
  % \vskip 0.2in
  \begin{center}
    \centerline{\includegraphics[width=\columnwidth]{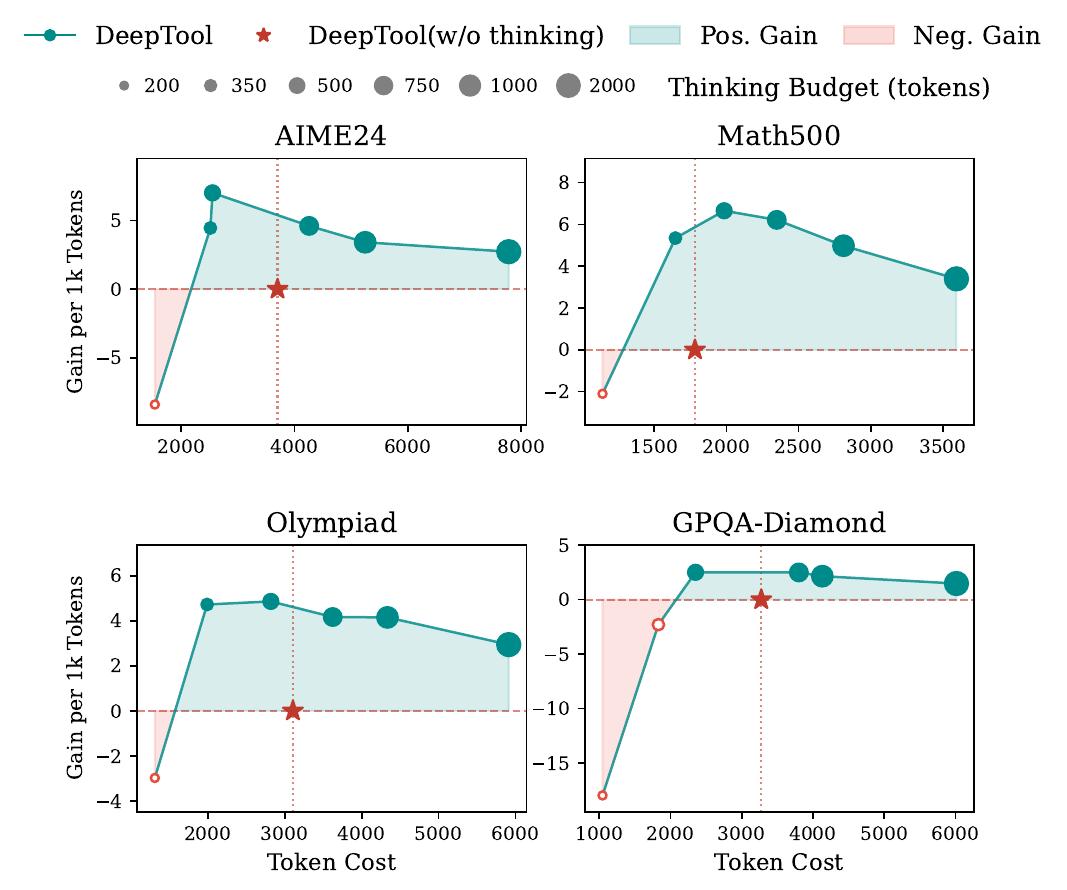}}
    \caption{
      Efficiency analysis of DeepTool. The y-axis reports the accuracy gain per 1,000 tokens relative to the non-thinking baseline (red star), computed by Eq.~\ref{eq:efficiency}. Shaded areas visualize cumulative net accuracy gain (blue) or loss (red) relative to the baseline. \textit{Peak} markers indicate the most cost-effective operating points.
    }
    \label{fig:efficiency}
  \end{center}
  \vskip -0.2in
\end{figure}

We evaluate the trade-off between performance improvement and computational overhead by analyzing the cost-effectiveness of Interleaved Thinking (Figure~\ref{fig:efficiency}).
Let $A_{\text{DeepTool}}$ and $A_{\text{base}}$ denote the accuracies of our method and the \textit{DeepTool (w/o thinking)} baseline, respectively. Let $C_{\text{DeepTool}}$ represent the average token cost per problem for our method. We define the cost-effectiveness metric (y-axis) as:
\begin{equation}
\label{eq:efficiency}
\mathrm{Eff} \;=\; \frac{A_{\text{DeepTool}} - A_{\text{base}}}{C_{\text{DeepTool}}}\times 1000,
\end{equation}
which measures the accuracy gain per $1\mathrm{k}$ tokens relative to the baseline. The sign and magnitude of $\mathrm{Eff}$ therefore capture token efficiency: larger values mean more accuracy improvement per token, while negative values indicate that additional tokens reduce accuracy. Full evaluation results across all tested benchmarks are provided in Appendix Figure~\ref{fig:efficiency_full}.

\paragraph{Reading the curves.}
The red star marks the baseline operating point, where $A_{\text{DeepTool}}=A_{\text{base}}$ and thus $\mathrm{Eff}=0$.
The shaded regions summarize the cumulative effect relative to the baseline: the blue area indicates accumulated net accurAacy gains, whereas the red area indicates accumulated net losses as compute increases.
Any configuration that lies in the \emph{upper-left} region with respect to the baseline marker (i.e., left of the red star with $\mathrm{Eff}>0$) achieves a strict improvement: it uses fewer tokens \emph{and} attains higher accuracy than direct tool execution, constituting a Pareto-dominant operating point.

\paragraph{Budget-driven efficiency dynamics.}
Across benchmarks, the curves exhibit a consistent pattern. In the low-budget regime, increasing the thinking budget raises $\mathrm{Eff}$, indicating that added reasoning space improves planning and verification, thereby increasing the accuracy return per token. 
As the budget continues to grow, $\mathrm{Eff}$ eventually declines: accuracy may still improve (reflected by the continued expansion of the positive shaded area), but the marginal gain per additional token diminishes.
Overall, the results demonstrate that optimal budget selection yields a favorable trade-off between token cost and accuracy, frequently surpassing baselines at reduced expense. Conversely, excessive budgets result in diminishing returns as performance saturates near the model's intrinsic limits.

%%%%%

%%%%%

\subsection{Analysis of Trajectory Synthesis Paradigms}
\label{sec:why_mosaic}

\begin{table}[t]
    \centering
    \caption{Performance comparison. \textit{Synthetic Trajectory Accuracy} denotes the percentage of trajectories yielding the correct final answer. Baseline refers to standard prompt-guided trajectory synthesis without guidance and adversarial perturbations. Datasets: \textbf{Aops} (Aops Forum), \textbf{CN} (CN Contest), \textbf{Ineq} (Inequalities), \textbf{NT} (Number Theory), \textbf{Olym} (Olympiads).}
    \label{tab:mosaic}

    \small % 使用 small 字体通常刚好填满单栏
    
    \begin{tabular}{lccccc} 
        \toprule
        \textbf{Method} & \textbf{Aops} & \textbf{CN} & \textbf{Ineq} & \textbf{NT} & \textbf{Olym} \\
        \midrule
        
        % --- Accuracy ---
        \multicolumn{6}{l}{\textit{\textbf{Synthetic Trajectory Accuracy (\%)}}} \\
        Mosaic   & \textbf{65.5} & \textbf{75.5} & \textbf{67.0} & \textbf{44.3} & \textbf{60.2} \\
        Baseline & 61.5          & 71.8          & 65.8          & 39.5          & 51.6          \\
        \midrule
        
        % --- Turns ---
        \multicolumn{6}{l}{\textit{\textbf{Avg. Turns}}} \\
        Mosaic   & \textbf{4.4} & \textbf{4.0} & \textbf{4.2} & \textbf{4.1} & \textbf{3.8} \\
        Baseline & 2.8          & 2.1          & 3.0          & 3.0          & 2.1          \\
        \midrule

        % --- Characteristics ---
        \multicolumn{6}{l}{\textit{\textbf{Paradigm}}} \\
        % 这里用了 multicolumn 跨越所有列，并将字体设为 footnotesize 以防撑破边界
        Mosaic   & \multicolumn{5}{l}{\footnotesize \textsc{Adv. Injection} $\oplus$ \textsc{Thinking}} \\
        Baseline & \multicolumn{5}{l}{\footnotesize \textit{Standard Generation}} \\
        
        \bottomrule
    \end{tabular}
\end{table}

We investigate the necessity of the MOSAIC pipeline by comparing it against a baseline approach, where training data is synthesized via standard prompting (detailed in Listing~\ref{lst:baseline-prompt}) that encourages multi-turn tool use without guidance and explicit error injection (Table~\ref{tab:mosaic}).

\paragraph{Stability and Robustness via Hierarchical Synthesis.} 
Standard prompt-guided synthesis lacks global planning, often succumbing to error propagation where a single logical deviation causes the entire trajectory to collapse. In contrast, MOSAIC significantly improves synthesis stability through the \textit{Manager Policy}, which enforces hierarchical constraints to keep the reasoning process aligned with the strategic intent, evidenced by consistently higher Synthetic Trajectory Accuracy (\textit{i.e.}, final answer correctness) across all datasets. Furthermore, by integrating \textit{Adversarial Injection}, MOSAIC actively exposes the model to execution anomalies during training. This combination ensures that the synthesized policy is not only structurally coherent but also robust, capable of rectifying errors through corrective reasoning rather than terminating upon failure.

\paragraph{Active Tool Engagement.}
The quantitative increase in average interaction turns reveals that MOSAIC fosters a more proactive reasoning strategy. Unlike baseline models that often converge prematurely or bypass necessary verification, MOSAIC leverages its hierarchical architecture to encourage extensive tool utilization. This higher interaction density reflects a learned behavior of \textit{iterative refinement}: the model actively decomposes complex logical leaps into executable sub-tasks and utilizes tool feedback for rigorous validation, thereby prioritizing solution correctness over superficial efficiency.

\section{Related Work}

\paragraph{System 2 Deliberation and Inference Scaling.}
The emergence of Chain-of-Thought prompting~\cite{wei2022chain} marked a paradigm shift in Large Language Models (LLMs), enabling them to decompose complex problems into intermediate steps~\cite{zhouleast}. The field has moved beyond fixed-computation paradigms toward test-time compute scaling, often termed System 2 reasoning~\cite{li2025system, xie2025interleaved}, where models allocate inference resources to deliberate before generating. More recently, large-scale reinforcement learning has unlocked models capable of "extended thinking," such as OpenAI's o1~\cite{jaech2024openai} series and DeepSeek-R1~\cite{guo2025deepseek}. However, these pure-language reasoning models remain prone to hallucination in computational tasks~\cite{farquhar2024detecting}, necessitating the integration of external tools~\cite{chenprogram}.

\paragraph{Tool-Integrated Reasoning (TIR).}
TIR emerged to enable LLMs to solve computationally intensive problems by leveraging external programming tools~\cite{chenprogram, song2025r1, wang2024empowering}. While \textit{Code-as-Reasoning} approaches like PAL~\cite{gao2023pal} and PoT~\cite{chenprogram} delegate reasoning entirely to executable programs, they sacrifice the flexibility of natural language abstraction. More recent frameworks, such as ReAct~\cite{yao2022react}, Toolformer~\cite{schick2023toolformer} and Search-R1~\cite{jin2025searchr}, interleave text generation with API calls. However, these methods typically rely on linear, greedy decision trajectories that are prone to error propagation and struggle with self-correction in complex multi-step settings~\cite{chen2025smurfs, liutool}. 
In contrast, DeepTool integrates System-2 deliberation into the tool-use loop, enabling dynamic verification and robust self-correction through interleaved thinking and execution.

% In contrast, DeepTool integrates System-2-level deliberation directly into the tool-use workflow. By tightly coupling deep thinking with tool execution at each step, DeepTool enables dynamic verification and revision during inference, supporting adaptive decision-making and robust error recovery.

\paragraph{Reinforcement Learning for TIR.}
Reinforcement Learning has advanced TIR by enabling autonomous exploration, utilizing verifiable outcome rewards to enhance generalization and optimize reasoning paths~\cite{li2025torl, feng2025retool, wei2025autotir}. However, these approaches suffer from the credit assignment problem in long-horizon tasks, where sparse rewards fail to distinguish between sound reasoning and spurious successful runs, and often neglecting the correctness of the \textit{thinking process} itself~\cite{lightman2023let, uesato2022solving}. While process supervision has proven effective for text-based reasoning~\cite{deng2025supervised}, its extension to multi-turn tool invocation remains unexplored. DeepTool bridges this gap by implementing Process-Supervised RL with an Action-Centric Process Reward. This approach provides dense step-wise supervision, resolving reward ambiguity and effectively fostering robust, adaptive \textit{System 2 interleaved thinking}.

% \paragraph{Reinforcement Learning for TIR.}
% Reinforcement Learning has advanced TIR by enabling autonomous exploration beyond supervised data. Prominent methods like ToRL~\cite{li2025torl} and ReTool~\cite{feng2025retool} employ outcome-supervised RL to optimize policies based solely on final answer correctness. However, these approaches suffer from the credit assignment problem in long-horizon tasks, where sparse rewards fail to distinguish between sound reasoning and spurious successful runs, and often neglecting the correctness of the \textit{thinking process} itself~\cite{lightman2023let, uesato2022solving}. While Supervised RL has proven effective for text-based reasoning via step-level supervision~\cite{deng2025supervised}, its extension to multi-turn tool invocation remains unexplored. Our framework implements Process-Supervised RL. By leveraging Group Relative Policy Optimization~\cite{shao2024deepseekmath} with our fine-grained Action-Centric Process Reward, we strictly validate both interleaved thinking and tool execution at each turn. This dense supervision resolves the ambiguity of sparse rewards, effectively fostering \textit{System 2 interleaved thinking}, a critical mechanism that drives adaptive strategic planning and robust self-correction throughout the reasoning process.

%%%%%%%%%%%%%%%%%%%%%%%%%%%%%%%%%%
\section{Conclusion}
\label{sec:conclusion}

In this paper, we propose \textit{\textbf{DeepTool}}, a novel framework that scales deliberate thinking within the interleaved process of thinking, action, and observation. To address shallow planning and sparse outcome-based rewards, we introduced two innovations: 
the \textbf{MOSAIC} synthesis pipeline, which incorporates interleaved deliberation and injects adversarial perturbations to cultivate intrinsic robustness and self-correction, 
and a \textbf{Process-Supervised Reinforcement Learning} strategy that leverages an \textit{Action-Centric Process Reward} to further scale interleaved deliberation and enforce precise tool invocations. Extensive experiments confirm \textit{\textbf{DeepTool}}'s superiority across challenging benchmarks. Moreover, our analysis validates the cost-effectiveness of ``interleaved deliberate thinking'', establishing a resilient paradigm that seamlessly unites cognitive depth with precise execution.

\section*{Impact Statement}

This paper presents work whose goal is to advance the field of Machine
Learning. There are many potential societal consequences of our work, none
which we feel must be specifically highlighted here.

\nocite{langley00}

\bibliography{references}
\bibliographystyle{icml2026}

%%%%%%%%%%%%%%%%%%%%%%%%%%%%%%%%%%%%%%%%%%%%%%%%%%%%%%%%%%%%%%%%%%%%%%%%%%%%%%%
%%%%%%%%%%%%%%%%%%%%%%%%%%%%%%%%%%%%%%%%%%%%%%%%%%%%%%%%%%%%%%%%%%%%%%%%%%%%%%%
% APPENDIX
%%%%%%%%%%%%%%%%%%%%%%%%%%%%%%%%%%%%%%%%%%%%%%%%%%%%%%%%%%%%%%%%%%%%%%%%%%%%%%%
%%%%%%%%%%%%%%%%%%%%%%%%%%%%%%%%%%%%%%%%%%%%%%%%%%%%%%%%%%%%%%%%%%%%%%%%%%%%%%%
\newpage
\appendix
\onecolumn

% \section{Limitations and Future Work} 
% \label{sec:limitations}

% While \textit{DeepTool} demonstrates significant advancements in tool-integrated reasoning, several limitations remain. First, the current framework relies on a structured schema to interleave deliberation and execution (i.e., the strict alternation of \texttt{<think>}, \texttt{<code>}, and \texttt{<observation>}). While this protocol stabilizes the reinforcement learning process, it imposes a fixed cadence that may not optimally reflect human-like problem solving, where the granularity of thinking versus acting varies dynamically based on task complexity. Consequently, the model has not yet achieved fully autonomous orchestration, where it can freely decide when to engage in prolonged deep thinking and when to execute rapid, consecutive tool invocations without predefined structural constraints.

% In future work, we aim to address this rigidity by developing mechanisms for fully dynamic deliberation-execution scheduling, allowing the model to autonomously determine the optimal density of thinking steps. Furthermore, while our current evaluation focuses on mathematical reasoning via code interpreters, we plan to extend the \textit{DeepTool} framework to heterogeneous tool environments. Integrating a broader spectrum of utilities (such as retrieval-augmented generation, web browsing, and domain-specific APIs) will be essential to validate the generalization of process-supervised interleaved thinking in complex, open-ended real-world scenarios.

\section{Experimental Details}

\subsection{Training Details}
\label{sec:impl_details}

We detail the training configurations below. DeepSeek-V3.2 is used for MOSAIC data synthesis, with experiments conducted on up to 4 NVIDIA A100 (80GB) GPUs. Table~\ref{tab:sft_hyperparams} lists the hyperparameters for both SFT variants.

Table~\ref{tab:rl_hyperparams} presents the Reinforcement Learning (GRPO) parameters. Equation~\ref{eq:grpo} generalizes the formulation for multiple updates ($\mu$, corresponding to \texttt{num\_iterations}) via a clipped surrogate objective. To prevent excessive deviation from the reference policy, $\text{clip}(\cdot, 1-\epsilon, 1+\epsilon)$ bounds the policy ratio within $[1-\epsilon, 1+\epsilon]$. For $\mu=1$, the objective simplifies to:

\begin{equation}
    \mathcal{L}_{\text{GRPO}}(\theta) = - \frac{1}{\sum_{i=1}^G |o_i|} \sum_{i=1}^G \sum_{k=1}^{|o_i|} \left[ \frac{\pi_\theta(o_{i,k} \mid q, o_{i,<k})}{[\pi_\theta(o_{i,k} \mid q, o_{i,<k})]_{\text{no grad}}} \hat{A}_{i,k} - \beta \mathbb{D}_{\text{KL}} [\pi_\theta || \pi_{\text{ref}}] \right]
\end{equation}

KL divergence is estimated using the approximator defined as follows:

\begin{equation}
\mathbb{D}_{\mathrm{KL}} [\pi_\theta \| \pi_{\mathrm{ref}}] = \frac{\pi_{\mathrm{ref}}(o_{i,k} \mid q, o_{i,<k})}{\pi_\theta(o_{i,k} \mid q, o_{i,<k})} - \log \frac{\pi_{\mathrm{ref}}(o_{i,k} \mid q, o_{i,<k})}{\pi_\theta(o_{i,k} \mid q, o_{i,<k})} - 1,
\end{equation}

\begin{table}[h]
\centering
\caption{Hyperparameters for the SFT stage. We compare the configurations between the \textit{w/o Thinking} and the \textit{Thinking} model.}
\label{tab:sft_hyperparams}
\begin{tabular}{l c c}
\toprule
\textbf{Hyperparameter} & \textbf{w/o Thinking SFT} & \textbf{Thinking SFT} \\
\midrule
Global Batch Size & 64 & 64 \\
Learning Rate & 1.0e-5 & 1.0e-5 \\
Total Epochs & 5 & 5 \\
LR Scheduler & Cosine & Cosine \\
Warmup Ratio & 0.1 & 0.1 \\
Mosaic probability $\rho$ & 0.1 & 0.1 \\
\midrule
\textit{Model-Specific Configs} & & \\
Max Sequence Length & 7,000 & 15,000 \\
DeepSpeed Strategy & ZeRO-2 & ZeRO-3 \\
Gradient Accumulation Steps & 32 & 16 \\
\bottomrule
\end{tabular}
\end{table}

\begin{table}[h]
\centering
\caption{Hyperparameters for the Reinforcement Learning (RL) stage using GRPO. This stage is initialized from the \textit{Thinking SFT} model.}
\label{tab:rl_hyperparams}
\begin{tabular}{l c}
\toprule
\textbf{Hyperparameter} & \textbf{Value} \\
\midrule
Algorithm & GRPO \\
Num Iterations & 1 \\
Global Batch Size & 32 \\
Learning Rate & 1.0e-6 \\
LR Scheduler & Linear \\
Warmup Ratio & 0.0 \\
Max Training Steps & 1,000 \\
\midrule
\textit{Generation \& Data} & \\
Mosaic probability $\rho$ & 0.2 \\
Num. Generations ($G$) & 4 \\
Temperature & 1.0 \\
Max Prompt Length & 16,000 \\
Max Completion Length & 6,000 \\
KL Coefficient ($\beta$) & 0.0 \\
Inference Engine & vLLM (Colocate) \\
\bottomrule
\end{tabular}
\end{table}

\newpage
\subsection{Complete Experimental Results}
\label{appendix:results}

This section provides the comprehensive data visualizations for the reasoning budget scaling laws and the computational efficiency analysis, supplementing the main text results.
% 1. Budget Scaling Full Plot
\begin{figure}[H] % h=here, t=top, b=bottom, p=page
    \centering
    % 使用 textwidth 占满版心宽度，keepaspectratio 保持长宽比
    \includegraphics[width=1.0\textwidth, keepaspectratio]{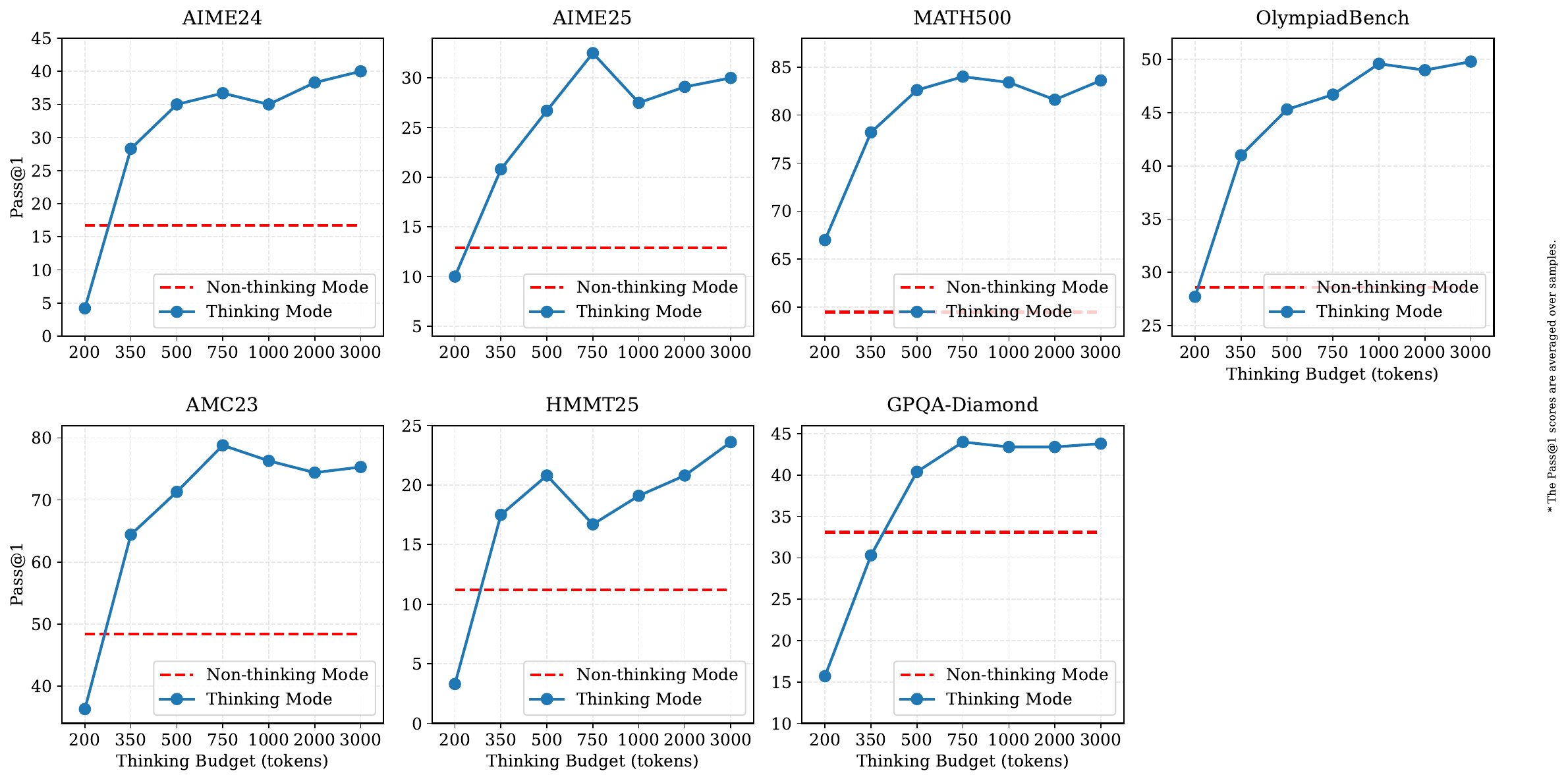}
    \caption{\textbf{Full Reasoning Budget Scaling Analysis.} 
    This figure illustrates the complete data spectrum regarding the correlation between thinking budget scaling and performance accuracy across varying difficulty levels. }
    \label{fig:budget_full}
\end{figure}

% 2. Efficiency Analysis Full Plot
\begin{figure}[H]
    \centering
    \includegraphics[width=1.0\textwidth, keepaspectratio]{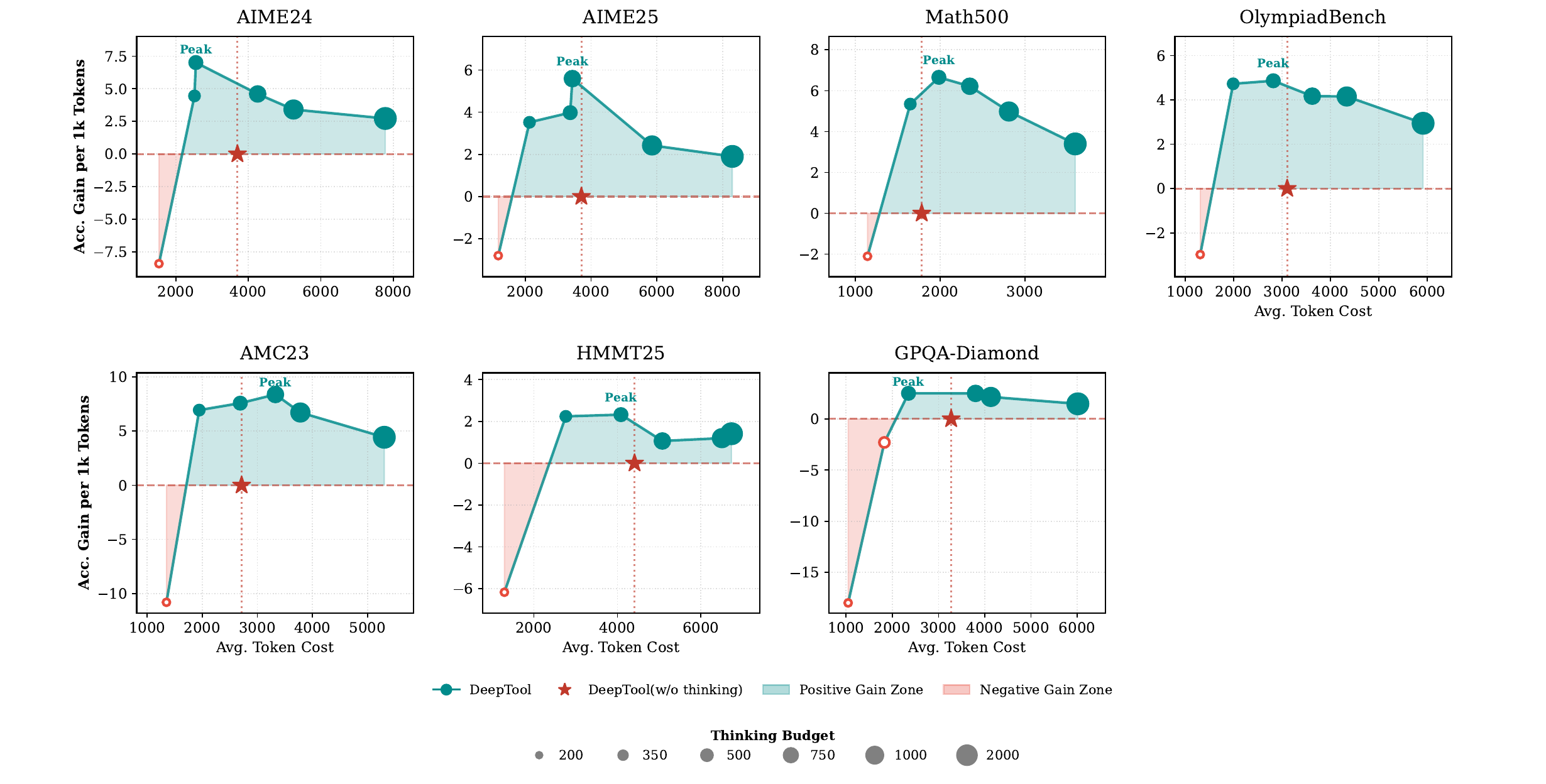}
    \caption{\textbf{Comprehensive Efficiency Analysis.} 
    Detailed breakdown of computational overhead versus performance gains. This visualization provides the full context for the efficiency analysis, highlighting the trade-offs between token consumption and reasoning depth.}
    \label{fig:efficiency_full}
    % \vskip -1in   
\end{figure}

\newpage
\subsection{Case Study}
\label{appendix:casestudy}

We present a detailed trajectory of DeepTool solving a motion problem using the Interleaved Tool-Integrated Reasoning framework. This demonstrates the interleaving of natural language reasoning, code generation, and execution feedback.

% 开始 Case Study 盒子
\begin{casestudybox}{Case Study: DeepTool}

% 问题描述
\stepheader{Problem Input}
\begin{quote}
The weight of $\frac{1}{3}$ of a large pizza together with $3 \frac{1}{2}$ cups of orange slices is the same as the weight of $\frac{3}{4}$ of a large pizza together with $\frac{1}{2}$ cup of orange slices. A cup of orange slices weighs $\frac{1}{4}$ of a pound. What is the weight, in pounds, of a large pizza? The answer can be written in the form $\frac{m}{n}$, where $m$ and $n$ are relatively prime positive integers. What is $m-n$?
\end{quote}

\stepheader{Gold Answer}

\begin{quote}
4
\end{quote}

\rule{\textwidth}{0.4pt} % 分割线

% --- Step 1 ---
\stepheader{Step 1}

\begin{thinkbox}
We are given: "The weight of 1/3 of a large pizza together with 3 1/2 cups of orange slices is the same as the weight of 3/4 of a large pizza together with 1/2 cup of orange slices. A cup of orange slices weighs 1/4 of a pound. What is the weight, in pounds, of a large pizza? The answer can be written in the form m/n, where m and n are relatively prime positive integers. What is m-n?"

We are to start with the Navigator's Guide: "Define a variable for the unknown weight of a large pizza. Then translate the given phrase comparing the weights into an algebraic equation, expressing the weight of orange slices in terms of cups and the given weight per cup."

So we will:
- Let P be the weight of a large pizza in pounds.
- Express the weight of orange slices: we have expressions involving cups and weight per cup. The weight of orange slices is volume in cups * weight per cup = cups * (1/4 lb per cup).
- The problem says: "The weight of 1/3 of a large pizza together with 3 1/2 cups of orange slices is the same as the weight of 3/4 of a large pizza together with 1/2 cup of orange slices."
So left side: (1/3) P + (3 1/2) * (1/4)
Right side: (3/4) P + (1/2) * (1/4)
Set them equal: (1/3)P + (1/2)*(3 1/2)*(1/4) = (3/4)P + (1/2)*(1/4)......
\end{thinkbox}

We define \( P \) as the weight of a large pizza in pounds. The weight of orange slices is cups multiplied by \( \frac{1}{4} \) lb per cup. The given condition translates to the equation:

\[
\frac{1}{3}P + \left(3\frac{1}{2}\right) \times \frac{1}{4} = \frac{3}{4}P + \left(\frac{1}{2}\right) \times \frac{1}{4}
\]

Let \( \text{orange} = \text{cups} \times \frac{1}{4} \)...

\begin{lstlisting}[language=Python, numbers=none]
from fractions imprt Fraction

weight_per_cup = Fraction(1, 4)          # lb per cup
orange_left = Fraction(7, 2)             # 3 1/2 cups = 7/2
orange_right = Fraction(1, 2)            # 1/2 cup

left_orange = orange_left * weight_per_cup  # 7/8
right_orange = orange_right * weight_per_cup # 1/8

A = Fraction(1, 3) - Fraction(3, 4)       # 1/3 - 3/4 = -5/12
B = right_orange - left_orange            # 1/8 - 7/8 = -6/8 = -3/4

P = B / A                                 # P = (-3/4) / (-5/12) = 9/5

print("Weight of large pizza:", P)
print(f"Fraction: {P.numerator}/{P.denominator}")
m = P.numerator
n = P.denominator
print("m - n =", m - n)
\end{lstlisting}

\begin{interpreterbox}
\texttt{Execution(Results: [], Logs: Logs(stdout: [], stderr: []), Error: ExecutionError(name='SyntaxError', value='invalid syntax (2603617310.py, line 6)', traceback='  Cell In[1], line 6  from fractions imprt Fraction SyntaxError: invalid syntax'))}
\end{interpreterbox}

% --- Step 2 ---
\stepheader{Step 2}

\begin{thinkbox}
We are given the problem and the previous steps. The previous code had a syntax error due to a misspelling of "import": "imprt" instead of "import". We need to correct this and also include the import statement. Then run the corrected code that computes P. According to the previous analysis, P = 9/5. The code will compute it and output m-n which is 4.

We'll write the corrected code.......
\end{thinkbox}

**Textual Analysis:**  
We define \( P \) as the weight of a large pizza in pounds. The weight of orange slices is given as cups multiplied by \( \frac{1}{4} \) lb per cup. The problem translates to:  
\[
\frac{1}{3}P + \frac{7}{2} \times \frac{1}{4} = \frac{3}{4}P + \frac{1}{2} \times \frac{1}{4}
\]

\begin{lstlisting}[language=Python, numbers=none]
from fractions import Fraction

weight_per_cup = Fraction(1, 4)          # lb per cup
orange_left = Fraction(7, 2)             # 3 1/2 cups = 7/2
orange_right = Fraction(1, 2)            # 1/2 cup

left_orange = orange_left * weight_per_cup  # 7/8
right_orange = orange_right * weight_per_cup # 1/8

A = Fraction(1, 3) - Fraction(3, 4)       # 1/3 - 3/4 = -5/12
B = right_orange - left_orange            # 1/8 - 7/8 = -6/8 = -3/4

P = B / A                                 # P = (-3/4) / (-5/12) = 9/5

print("Weight of large pizza:", P)
print(f"Fraction: {P.numerator}/{P.denominator}")
m = P.numerator
n = P.denominator
print("m - n =", m - n)
\end{lstlisting}

\begin{interpreterbox}
Weight of large pizza: 9/5
Fraction: 9/5
m - n = 4
\end{interpreterbox}

% --- Step 3 ---
\stepheader{Step 3}

\begin{thinkbox}
We are given the problem and the history. The problem asks for the weight of a large pizza in pounds, expressed as a fraction m/n in simplest form, and then m-n. We have gone through the steps: defining the variable, setting up the equation, solving, and obtaining P = 9/5. Thus m=9, n=5, and m-n = 4.

We summarize the reasoning: Let P be the weight of a large pizza. The weight of orange slices is cups times 0.25 lb per cup. Setting the two combined weights equal gives the equation (1/3)P + (7/2)*(1/4) = (3/4)P + (1/2)*(1/4). Solving yields P = 9/5. Therefore m=9, n=5, and m-n = 4.

Now present the final answer.
\end{thinkbox}

Let \(P\) be the weight (in pounds) of a large pizza. The orange slices weigh \(0.25\) lb per cup, so the total weight on the left is \(\frac{1}{3}P + \frac{7}{2} \times \frac{1}{4}\) and on the right is \(\frac{3}{4}P + \frac{1}{2} \times \frac{1}{4}\). Setting these equal and solving gives \(P = \frac{9}{5}\). Thus, \(m = 9\) and \(n = 5\), so \(m - n = 4\).

\vspace{1em}
\noindent\textbf{Final Answer:}
\[ \boxed{4} \]

\end{casestudybox}

%%%%%%%%%%%%%%%%%%%
\newpage

% 开始 Case Study 盒子
\begin{casestudybox}{Case Study: w/o thinking TIR}

% 问题描述
\stepheader{Problem Input}
\begin{quote}
The weight of $\frac{1}{3}$ of a large pizza together with $3 \frac{1}{2}$ cups of orange slices is the same as the weight of $\frac{3}{4}$ of a large pizza together with $\frac{1}{2}$ cup of orange slices. A cup of orange slices weighs $\frac{1}{4}$ of a pound. What is the weight, in pounds, of a large pizza? The answer can be written in the form $\frac{m}{n}$, where $m$ and $n$ are relatively prime positive integers. What is $m-n$?
\end{quote}

\stepheader{Gold Answer}

\begin{quote}
4
\end{quote}

\rule{\textwidth}{0.4pt} % 分割线

% --- Step 1 ---
\stepheader{Step 1}

First, express all weights. Let \( p \) be the weight of a large pizza in pounds. A cup of orange slices weighs \( \frac{1}{4} \) lb. The total weight for each side is computed exactly using fractions.

\begin{lstlisting}[language=Python, numbers=none]
from fractions import Fraction

# Given values
cup_weight = Fraction(1, 4)   # weight of one cup of orange slices (lb)
orange_3_1_2 = Fraction(7, 2) # 3 1/2 cups = 7/2 cups

# Express the two sides
left_side = Fraction(1, 3) * Fraction('p') + orange_3_1_2   # (1/3)p + 7/2
right_side = Fraction(3, 4) * Fraction('p') + Fraction(1, 2) # (3/4)p + 1/2

# Print in exact fraction form
print("Left side total weight (pounds):")
print(f"({Fraction(1,3)}) * p + {orange_3_1_2}")
print(f"Which simplifies to: ({Fraction(1,3)} * {orange_3_1_2}) + p = {Fraction(1,3)*orange_3_1_2} + p")
print(f"Or: p + {Fraction(1,3)*orange_3_1_2}")
print()
print("Right side total weight (pounds):")
print(f"({Fraction(3,4)}) * p + {Fraction(1,2)}")
print(f"Which simplifies to: ({Fraction(3,4)} * {Fraction(1,2)}) + p = {Fraction(3,4)*Fraction(1,2)} + p")
print(f"Or: p + {Fraction(3,4)*Fraction(1,2)}")
\end{lstlisting}

\begin{interpreterbox}
Execution(Results: [], Logs: Logs(stdout: [], stderr: []), Error: ExecutionError(name='ValueError', value="Invalid literal for Fraction: 'p'", traceback="---------------------------------------------------------------------------ValueError                                Traceback (most recent call last)Cell In[1], line 13\\n     10 orange = Fraction(7, 2) \# 3 1/2 cups = 7/2 cups\\n     12 \# Express the two sides\\n---> 13 left side = Fraction(1, 3) * Fraction('p') + orange   \# (1/3)p + 7/2\\n     14 right side = Fraction(3, 4) * Fraction('p') + Fraction(1, 2) \# (3/4)p + 1/2\\n     16 \# Print in exact fraction form\\nFile /usr/local/lib/python3.12/fractions.py:239, in Fraction. new (cls, numerator, denominator)\\n    237 m =  RATIONAL FORMAT.match(numerator)\\n    238 if m is None:\\n--> 239     raise ValueError('Invalid literal for Fraction: 240                      numerator)\\n    241 numerator = int(m.group('num') or '0')\\n    242 denom = m.group('denom')\\nValueError: Invalid literal for Fraction: 'p'"))
\end{interpreterbox}

% --- Step 2 ---
\stepheader{Step 2}

**Textual Analysis:** The previous code attempted to compute `Fraction(1,3)*Fraction('p')`, which is invalid because `Fraction` expects integer arguments. To set up the equation, we treat `p` as a symbolic variable. We will define a function that computes the total weights for both sides given `p` and then equate them. Alternatively, we can write the equation as `left side = right side` without attempting arithmetic operations involving the unknown `p`. This avoids the error and prepares for solving the equation in the next step.

\begin{lstlisting}[language=Python, numbers=none]
from fractions import Fraction

# Given values
cup_weight = Fraction(1, 4)       # weight of one cup of orange slices (lb)
orange_3_1_2 = Fraction(7, 2)     # 3 1/2 cups = 7/2 cups

# Express the two sides symbolically
left_side_expr = "({}/3) * p + {}".format(orange_3_1_2.numerator, orange_3_1_2.denominator)
right_side_expr = "({}/4) * p + {}".format(Fraction(1,2).numerator, Fraction(1,2).denominator)

print("Derived equation from the problem statement:")
print(f"Left side = Right side")
print(f"({orange_3_1_2} cups of orange slices weigh {orange_3_1_2 * cup_weight} lb)")
print(f"Equation: ({left_side_expr} + 7/4) = (3/4 * p + 1/2)")
print(f"Simplified: ({orange_3_1_2} * 1/4) + p = (3/4)p + 1/2")
print(f"i.e., {orange_3_1_2 / 4} + p = (3/4)p + 1/2")
print(f"Or, in fractional terms: p + {orange_3_1_2}p = 1/2 + 7/4")
print(f"Equation to solve: p + {orange_3_1_2}p - 1/2 - 7/4 = 0")
print(f"Combined: p(1 + {orange_3_1_2}) - (1/2 + 7/4) = 0")
print(f"Thus: p * (4/3 + 7/2) = 1/2 + 7/4")
print(f"Or: p * (1/3 + 7/2) = ...")
\end{lstlisting}

\begin{interpreterbox}
Derived equation from the problem statement:
Left side = Right side
(7/2 cups of orange slices weigh 7/8 lb)
Equation: ((7/3) * p + 2 + 7/4) = (3/4 * p + 1/2)
Simplified: (7/2 * 1/4) + p = (3/4)p + 1/2
i.e., 7/8 + p = (3/4)p + 1/2
Or, in fractional terms: p + 7/2p = 1/2 + 7/4
Equation to solve: p + 7/2p - 1/2 - 7/4 = 0
Combined: p(1 + 7/2) - (1/2 + 7/4) = 0
Thus: p * (4/3 + 7/2) = 1/2 + 7/4
Or: p * (1/3 + 7/2) = ...
\end{interpreterbox}

% --- Step 3 ---
\stepheader{Step 3}

**Textual Analysis:** We equate the two total weight expressions and solve for \( p \), the weight of a large pizza in pounds. Algebraically, we have:
\[
\left(\frac{1}{3}\right)p + \frac{7}{2} \cdot \frac{1}{4} = \left(\frac{3}{4}\right)p + \frac{1}{2}
\]
which simplifies to:
\[
\frac{7}{8} + \frac{4}{3}p = \frac{1}{2} + \frac{7}{4}
\]
or:
\[
\frac{7}{8} + \frac{4}{3}p = \frac{9}{4}
\]
Then:
\[
\frac{4}{3}p = \frac{9}{4} - \frac{7}{8} = \frac{11}{8}
\]
Thus:
\[
p = \frac{11}{8} \div \frac{4}{3} = \frac{11}{8} \cdot \frac{3}{4} = \frac{33}{32}
\]
The code below verifies this by collecting like terms and applying fraction arithmetic.

\begin{lstlisting}[language=Python, numbers=none]
from fractions import Fraction

# Define Fraction objects for all numbers
left_coeff = Fraction(1, 3)          # coefficient of p on left
right_coeff = Fraction(3, 4)         # coefficient of p on right
orange_amount = Fraction(7, 2)       # 3 1/2 cups
cup_weight = Fraction(1, 4)          # weight per cup
constant_right = Fraction(1, 2)      # constant on right

# Compute constant on left: orange weight
orange_weight = orange_amount * cup_weight

# Equation: left_coeff * p + orange_weight = right_coeff * p + constant_right
# Rearranged: (left_coeff - right_coeff) * p = constant_right - orange_weight
coeff_p = left_coeff - right_coeff
constant = constant_right - orange_weight

# Solve for p: p = constant / coeff_p
p_solution = constant / coeff_p

# Output in a nice form
print("Equation setup:")
print(f"({left_coeff})*p + ({orange_weight}) = ({right_coeff})*p + {constant_right}")
print()
print("Simplifying:")
print(f"left-hand side constant: {orange_weight}")
print(f"right-hand side constant: {constant}")
print(f"Then: p * ({coeff_p}) = {constant}")
print(f"Thus: p = {constant} / {coeff_p}")
print()
print("Computed weight of a large pizza:")
print(f"p = {p_solution} pounds (exactly)")
print(f"   = {float(p_solution):.4f} lb")
print()
# Show as fraction in simplest form (already simplified by Fraction)
if p_solution.numerator != 0:
    gcd = Fraction.gcd(p_solution.numerator, p_solution.denominator)
    print(f"Fraction in lowest terms: {p_solution.numerator // gcd}/{p_solution.denominator // gcd}")
else:
    print("p = 0 (no pizza)")
\end{lstlisting}

\begin{interpreterbox}
Equation setup:
(1/3)*p + (7/8) = (3/4)*p + 1/2

Simplifying:
left-hand side constant: 7/8
right-hand side constant: -3/8
Then: p * (-5/12) = -3/8
Thus: p = -3/8 / -5/12

Computed weight of a large pizza:
p = 9/10 pounds (exactly)
   = 0.9000 lb
\end{interpreterbox}

\textbf{......}

\stepheader{Step 5}

In the previous step, we attempted to use `Fraction('p')` to denote the unknown variable \( p \), which caused a `ValueError` because `Fraction` expects numeric literals. The Navigator’s Guide corrects this by solving the equation algebraically without symbolic representation. Starting from the equation \( \frac{p}{3} + \frac{7}{8} = \frac{3p}{4} + \frac{1}{2} \), we rearrange to isolate \( p \): move all \( p \)-terms to one side and constants to the other...

\begin{lstlisting}[language=Python, numbers=none]
from fractions import Fraction

# Constants from the equation: p/3 + 7/8 = 3p/4 + 1/2
constant_left = Fraction(7, 8)          # orange slices weight
constant_right = Fraction(1, 2)         # constant on right side
coeff_p_left = Fraction(1, 3)           # coefficient of p on left
coeff_p_right = Fraction(3, 4)          # coefficient of p on right

# Rearranged: p/3 - 3p/4 = 1/2 - 7/8
left_coeff = coeff_p_left - coeff_p_right   # 1/3 - 3/4 = -5/12
right_constant = constant_right - constant_left  # 1/2 - 7/8 = -3/8

print("Step-by-step algebraic solution using exact fractions:")
print(f"Equation: ({coeff_p_left})p + {constant_left} = ({coeff_p_right})p + {constant_right}")
print()
print("1. Move p-terms to left, constants to right:")
print(f"   {constant_left} - {constant_right} = ({coeff_p_right})p - ({coeff_p_left})p")
print()
print(f"2. Compute left constant difference: {constant_left} - {constant_right} = {right_constant}")
print(f"3. Compute right coefficient difference: ({coeff_p_right})p - ({coeff_p_left})p = {left_coeff}p")
print()
print(f"4. Resulting equation: {right_constant} = {left_coeff} * p")
print(f"5. Solve for p: p = {right_constant} / {left_coeff}")
print()
p_solution = right_constant / left_coeff
print(f"6. p = {p_solution} pounds")
print(f"   Reduced fraction: {p_solution.numerator}/{p_solution.denominator}")
print()
print("Verification:")
ver1 = coeff_p_left * p_solution + constant_left
ver2 = coeff_p_right * p_solution + constant_right
print(f"  Left side: ({coeff_p_left})*{p_solution} + {constant_left} = {ver1}")
print(f"  Right side: ({coeff_p_right})*{p_solution} + {constant_right} = {ver2}")
print(f"  {'='*30}")
print(f"  {'OK' if ver1 == ver2 else 'MISMATCH' }")
\end{lstlisting}

\begin{interpreterbox}
Step-by-step algebraic solution using exact fractions:
Equation: (1/3)p + 7/8 = (3/4)p + 1/2

1. Move p-terms to left, constants to right:
   7/8 - 1/2 = (3/4)p - (1/3)p

2. Compute left constant difference: 7/8 - 1/2 = -3/8
3. Compute right coefficient difference: (3/4)p - (1/3)p = -5/12p

4. Resulting equation: -3/8 = -5/12 * p
5. Solve for p: p = -3/8 / -5/12

6. p = 9/10 pounds
   Reduced fraction: 9/10

Verification:
  Left side: (1/3)*9/10 + 7/8 = 47/40
  Right side: (3/4)*9/10 + 1/2 = 47/40
  ==============================
  OK
\end{interpreterbox}

\stepheader{Step 6}

**Summary Analysis:**  
The algebraic solution to the equation derived from the problem yields the weight of a large pizza as \( \frac{9}{10} \) pounds. Here, \( m = 9 \) and \( n = 10 \), which are relatively prime. The problem asks for \( m - n \), which is \( 9 - 10 = -1 \).

\vspace{1em}
\noindent\textbf{Final Answer:}
\[ \boxed{-1} \]

\end{casestudybox}

%%%%%%%%%%%%%%%%%%%%%%%%%%%%%%%%%%%%%%%%%%%%%%%%%%%%%%%%%%%%%%%%%%%%%%%%%%%%%%%
%%%%%%%%%%%%%%%%%%%%%%%%%%%%%%%%%%%%%%%%%%%%%%%%%%%%%%%%%%%%%%%%%%%%%%%%%%%%%%%

\subsection{Prompts for DeepTool}
\label{appendix:prompts}

The following sections detail the specific prompt templates used in our DeepTool.

\begin{lstlisting}[caption={Prompt for Manager Policy}, numbers=none]
def thinking_navigation_card_generation_prompt(
    problem: str,
    gold_cot: str,
    history: str,
    remaining_steps: int
):
    # Handle empty history
    if not history or history.strip() == "":
        history = "No previous steps taken. This is the very first step."

    # Urgency control logic
    if remaining_steps <= 1:
        urgency_instruction = (
            "CRITICAL WARNING: You have ONLY 1 step remaining. "
            "You MUST instruct the Assistant to conclude all reasoning "
            "and generate the Final Answer in this step."
        )
    else:
        urgency_instruction = (
            f"You have {remaining_steps} steps remaining. "
            "Plan accordingly to ensure completion within the limit."
        )

    return f"""
[Role]
You are the "Thinking Navigator" in a step-by-step reasoning process.
You have access to the Problem, the Gold CoT, and Conversation History.
Your goal is to generate a navigation card for the NEXT step.

[Input Data]
1. Problem: \"\"\"{problem}\"\"\"
2. Gold CoT: \"\"\"{gold_cot}\"\"\"
3. History: \"\"\"{history}\"\"\"

[Constraint]
{urgency_instruction}

[Guiding Principles]
- Process over Result: Define the logical task, DO NOT reveal numeric results.
- Let the Assistant Think: The Assistant performs the execution.
- Handling Errors: If previous step failed, guide the fix first.

[Analysis Logic]
1. Check for Execution Errors (Highest Priority):
   - Look for <interpreter> error messages.
   - Action: Set reasoning_mode="Correction", guide debugging.
2. Check for Logical Validity:
   - Compare reasoning against Gold CoT.
   - If invalid: Set reasoning_mode="Correction".
   - If valid: Set reasoning_mode="Progress".
3. Determine Stage:
   - "Correction" -> "Intermediate"
   - Last step or Complete -> "Final"
   - Otherwise -> "Intermediate"

[Output Format]
JSON Object only:
{{
  "navigational_guide": "Concise guideline for the NEXT step...",
  "solution_stage": "Intermediate" | "Final",
  "reasoning_mode": "Progress" | "Correction"
}}
"""
\end{lstlisting}

\begin{lstlisting}[caption={Prompt for the Actor Policy}, numbers=none]
def step_thinking_tir_prompt(
    problem: str,
    history: str,
    thinking_navigation_card  
):
    # 1. Handle History
    history_str = history if history.strip() else "No previous steps."

    # 2. Extract info
    guide = thinking_navigation_card.navigational_guide
    stage = thinking_navigation_card.solution_stage
    mode = thinking_navigation_card.reasoning_mode

    # 3. Context Instruction
    if mode == "Correction":
        mode_context = (
            "\n[ATTENTION: CORRECTION MODE]\n"
            "Navigator identified an error. Explain what was wrong and how to fix it."
        )
    else:
        mode_context = "\n[MODE: PROGRESS]\nProceed logically based on the guide."

    # 4. Format Requirements
    if stage == "Final":
        task_instr = "Synthesize history and DERIVE THE FINAL ANSWER."
        fmt_req = """
**Output Format:**
1. Summary Analysis.
2. <answer>\\boxed{Final Answer}</answer>
"""
    else:  # Intermediate
        task_instr = "WRITE PYTHON CODE to perform the task."
        fmt_req = """
**Output Format:**
1. Textual Analysis (Logic/Math).
2. <code>
```python
# Code here

```

</code>
"""

```
    return f"""You are an expert solver using "Tool-Integrated Reasoning".

```

You are the "Builder" executing the plan provided by the "Navigator".

[Input Data]
Problem: """{problem}"""
History: """{history_str}"""
Navigator's Guide: """{guide}"""

[Task Context]
{mode_context}

[Goal]
{task_instr}

[Requirements]
Follow the guide exactly. {fmt_req}
"""
\end{lstlisting}

\begin{lstlisting}[caption={System Prompt for DeepTool Inference}, label={lst:DeepTool-Infer}, numbers=none]
def infer_prompt(problem: str) -> str:
    return f"""You are a helpful assistant solving math problems with Python.

[Protocol]

1. Step-by-Step Reasoning: Do not rush.
2. Tool Usage: Wrap code in <code> blocks. Analyze output.
3. Error Handling: Fix syntax/timeout errors in the next step.
4. Final Answer: Wrap result in <answer> block.

[Format Definitions]
Option 1: Code Step
<think>...</think>
Analysis...
<code>`python...`</code>

Option 2: Answer Step
<think>...</think>
Summary...
<answer>\boxed{{final_value}}</answer>

[Problem]
{problem}
"""
\end{lstlisting}

\begin{lstlisting}[caption={Prompt for Baseline Trajectory Synthesis Method}, label={lst:baseline-prompt}, numbers=none]
def generate_baseline_prompt(question: str) -> str:
    return f"""Solve the following problem step by step. You can write Python code.
The code output (in <interpreter>) will aid your reasoning.

Code Format:
<code>

```python
code snippet

```

</code>

Answer Format: <answer>\boxed{{'answer'}}</answer>

User Question: {question}

Assistant:"""
\end{lstlisting}

\end{document}